%% file: main.tex
\documentclass[10pt,twocolumn,letterpaper]{article}
\usepackage[accsupp]{axessibility} 
\usepackage[pagenumbers]{cvpr} 

\input{preamble}

\usepackage{bm}
\usepackage{bbm}
\usepackage{svg}
\usepackage{xcolor}
\usepackage{pifont}
\usepackage{comment}
\usepackage{booktabs}
\usepackage{multirow}
\usepackage{adjustbox}
\usepackage{mathtools}
\usepackage{tablefootnote}
\usepackage{threeparttable}

\newcommand{\cmark}{\ding{51}}

\definecolor{cvprblue}{rgb}{0.21,0.49,0.74}
\usepackage[pagebackref,breaklinks,colorlinks,citecolor=cvprblue]{hyperref}

\def\paperID{8994} 
\def\confName{CVPR}
\def\confYear{2024}

\title{VSRD: Instance-Aware Volumetric Silhouette Rendering\\ for Weakly Supervised 3D Object Detection}

\author{
Zihua Liu$^{*}$ \\
Tokyo Institute of Technology \\
{\tt\small zliu@ok.sc.e.titech.ac.jp}
\and
Hiroki Sakuma$^{*}$ \\
T2 Inc. \\
{\tt\small sakuma.h@t2.auto}
\and
Masatoshi Okutomi \\
Tokyo Institute of Technology \\
{\tt\small mxo@sc.e.titech.ac.jp}
}

\begin{document}
\maketitle

\def\thefootnote{*}\footnotetext{Equal contribution. The order was determined by a coin flip.}\def\thefootnote{\arabic{footnote}}

\input{abstract/main}    
\input{introduction/main}
\input{related_work/main}

\input{method/main}

\input{experiments/main}

\input{conclusion/main}

\clearpage

{
    \small
    \bibliographystyle{ieeenat_fullname}
    \bibliography{main}
}


\clearpage
\setcounter{page}{1}
\maketitlesupplementary

\input{additional_implementation_details/main}

\input{additional_evaluation_results/main}

\input{additional_visualization_results/main}

\end{document}

%% file: preamble.tex
\usepackage[dvipsnames]{xcolor}


%% file: abstract/main.tex
\begin{abstract}
Monocular 3D object detection poses a significant challenge in 3D scene understanding due to its inherently ill-posed nature in monocular depth estimation. Existing methods heavily rely on supervised learning using abundant 3D labels, typically obtained through expensive and labor-intensive annotation on LiDAR point clouds. To tackle this problem, we propose a novel weakly supervised 3D object detection framework named VSRD (\underline{\textbf{V}}olumetric \underline{\textbf{S}}ilhouette \underline{\textbf{R}}endering for \underline{\textbf{D}}etection) to train 3D object detectors without any 3D supervision but only weak 2D supervision. VSRD consists of multi-view 3D auto-labeling and subsequent training of monocular 3D object detectors using the pseudo labels generated in the auto-labeling stage. In the auto-labeling stage, we represent the surface of each instance as a signed distance field (SDF) and render its silhouette as an instance mask through our proposed instance-aware volumetric silhouette rendering. To directly optimize the 3D bounding boxes through rendering, we decompose the SDF of each instance into the SDF of a cuboid and the residual distance field (RDF) that represents the residual from the cuboid. This mechanism enables us to optimize the 3D bounding boxes in an end-to-end manner by comparing the rendered instance masks with the ground truth instance masks. The optimized 3D bounding boxes serve as effective training data for 3D object detection. We conduct extensive experiments on the KITTI-360 dataset, demonstrating that our method outperforms the existing weakly supervised 3D object detection methods. The code is available at \url{https://github.com/skmhrk1209/VSRD}.
\end{abstract}

%% file: introduction/main.tex
\input{method/figure}

\section{Introduction}
\label{sec:introduction}

3D object detection is one of the most critical components in the perception system for autonomous driving. With the recent success of deep learning in computer vision, numerous 3D object detection methods have been proposed to directly regress 3D bounding boxes from a LiDAR point cloud, multi-view images, or a monocular image. Among them, monocular 3D object detection is the most challenging in principle due to its inherently ill-posed nature in monocular depth estimation. Therefore, existing methods \cite{FCOS3D, DD3D, MonoDIS, MonoFlex, MonoDETR} heavily rely on supervised learning using abundant 3D labels manually annotated on LiDAR point clouds. This annotation cost is extremely high, posing a significant barrier to deploying 3D object detectors into autonomous driving systems. 

To tackle this problem, we propose a novel weakly supervised 3D object detection framework named VSRD (\underline{\textbf{V}}olumetric \underline{\textbf{S}}ilhouette \underline{\textbf{R}}endering for \underline{\textbf{D}}etection) to train 3D object detectors without any 3D supervision but only weak 2D supervision. As illustrated in \cref{fig:method}, VSRD consists of multi-view 3D auto-labeling and subsequent training of monocular 3D object detectors using the pseudo labels generated in the auto-labeling stage. In the auto-labeling stage, we represent the surface of each 3D bounding box as a signed distance field (SDF) and render its silhouette as an instance mask through volumetric rendering. Comparing the rendered instance masks with the ground truth instance masks enables us to optimize the 3D bounding boxes directly. We introduce two novel mechanisms in this auto-labeling stage. The first is the instance-aware volumetric silhouette rendering that integrates instance labels along a ray to render the silhouette of each instance rather than the entire scene. This mechanism enables the silhouette of each instance to be rendered while considering geometric relationships among instances, such as occlusion. The second is the SDF decomposition, whereby the SDF of each instance is decomposed into the SDF of a cuboid and the \textit{residual distance field} (RDF) that represents the residual from the cuboid. This decomposition models the spatial gap between the surfaces of each instance and the 3D bounding box, enabling more accurate silhouette rendering and providing more reliable feedback signals during optimization.

The 3D bounding boxes optimized by the proposed auto-labeling can serve as pseudo labels for training 3D object detectors. However, dynamic objects and inaccurate camera poses lead to low-quality pseudo labels, which negatively impact the training of 3D object detectors. Therefore, we propose a simple but effective algorithm to assign a confidence score representing the label quality to each pseudo label. We demonstrate that using these confidence scores as per-instance loss weights boosts the performance of the 3D object detectors trained on the pseudo labels. 

In summary, our main contributions are as follows:
\begin{itemize}
  \item We propose a weakly supervised 3D object detection framework consisting of multi-view 3D auto-labeling and subsequent training of monocular 3D object detectors.
  \item We propose a novel instance-aware volumetric silhouette rendering method whereby the silhouette of each instance can be rendered as an instance mask. 
  \item We propose a novel SDF decomposition whereby the SDF of each instance is decomposed into the SDF of a cuboid and the residual distance field (RDF). 
  \item We propose a simple but effective confidence assignment algorithm to incorporate the quality of each pseudo label into the training of 3D object detectors.
  \item Extensive experiments on the KITTI-360 dataset demonstrate that our method outperforms the existing weakly supervised 3D object detection methods.
\end{itemize}

%% file: method/figure.tex
\begin{figure}[t]
    \centering
    \includegraphics[width=\linewidth]{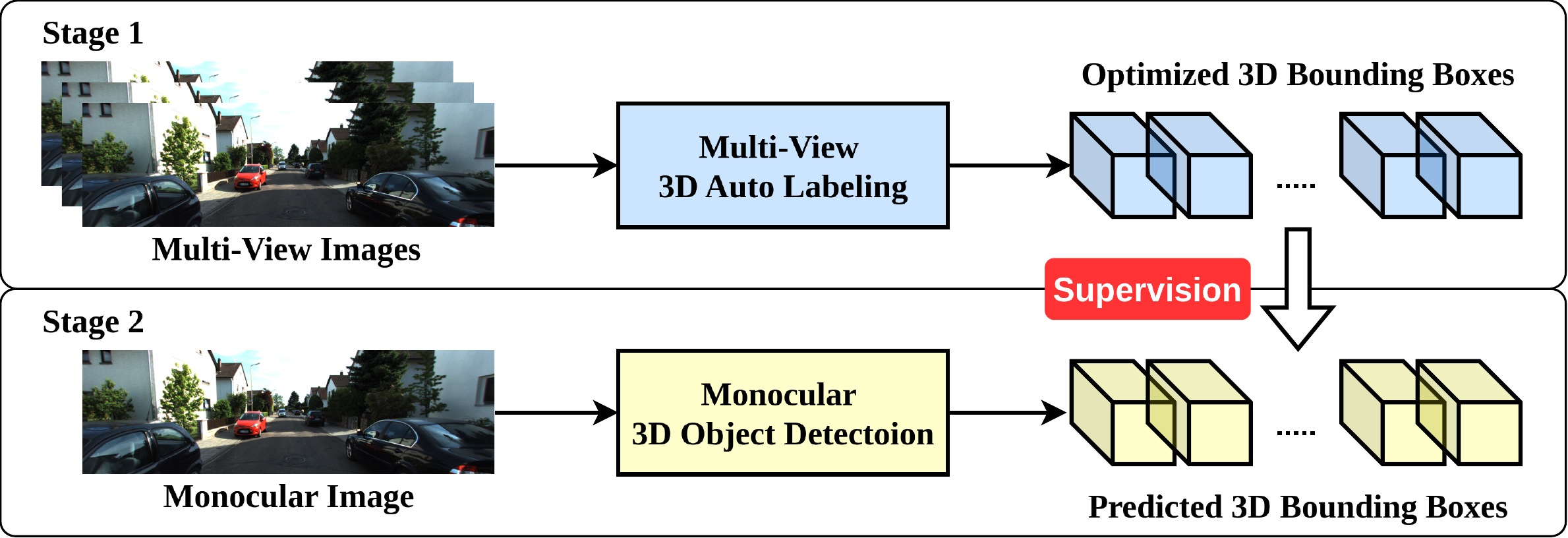}
    \caption{Illustration of our proposed weakly supervised 3D object detection framework, which consists of multi-view 3D auto-labeling and subsequent training of monocular 3D object detectors using the pseudo labels generated in the auto-labeling stage.}
    \label{fig:method}
\end{figure}

%% file: related_work/main.tex
\input{method/multi_view_3d_auto_labeling/figure}

\section{Related Work}
\label{sec:related_work}

\input{related_work/monocular_3d_object_detection/main}
\input{related_work/weakly_supervised_3d_object_detection/main}
\input{related_work/3d_object_detection_with_neural_fields/main}

%% file: method/multi_view_3d_auto_labeling/figure.tex
\begin{figure*}[t]
    \centering
    \includegraphics[width=\linewidth]{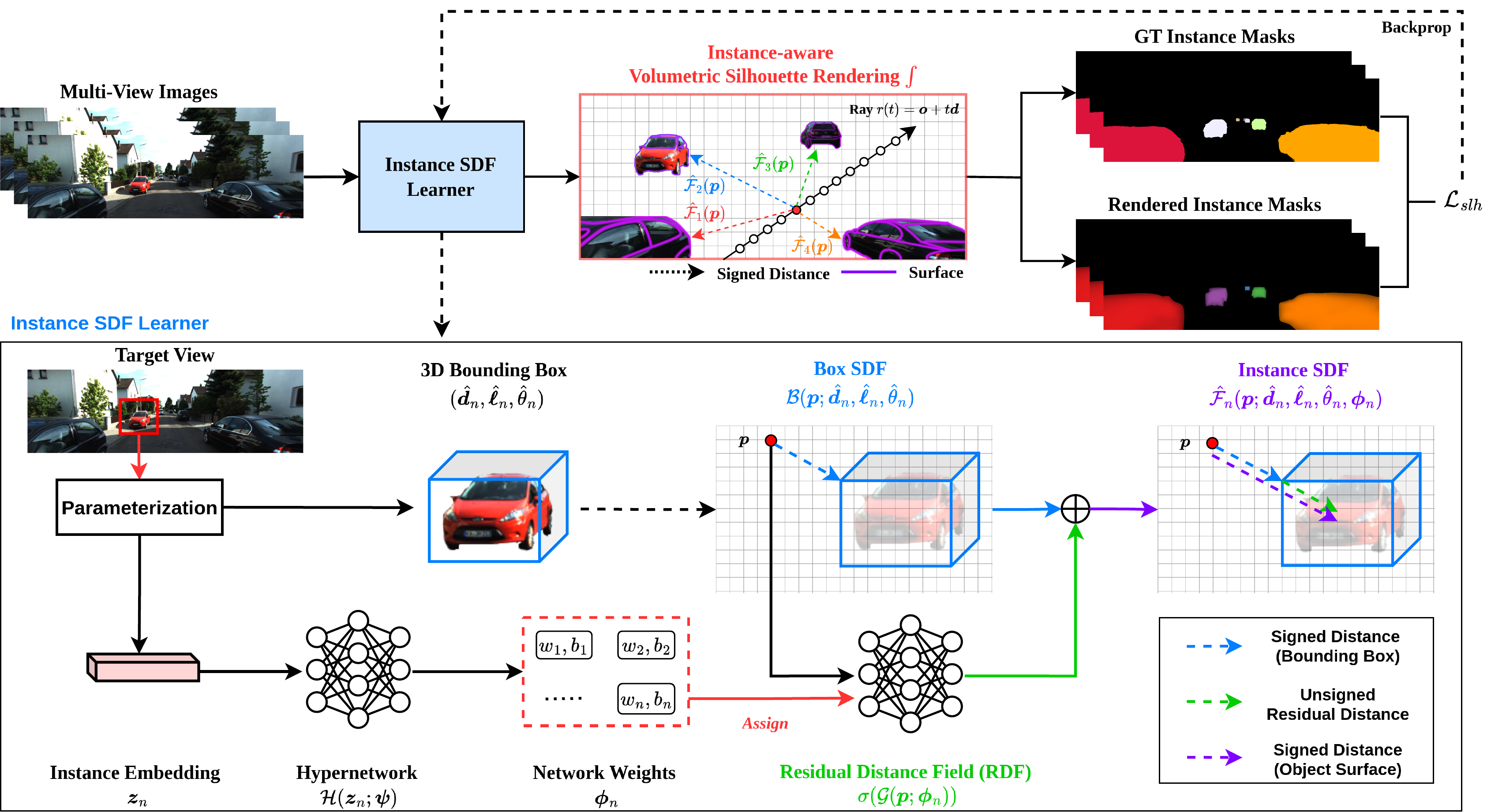}
    \caption{Illustration of the pipeline of our proposed multi-view 3D auto-labeling. We represent the surface of each instance as an SDF and decompose it into the SDF of a 3D bounding box and the \textit{residual distance field} (RDF), which is learned via a hypernetwork. The composed instance SDF is used to render the silhouette of the instance through our proposed instance-aware volumetric silhouette rendering. All the 3D bounding boxes are optimized based on the loss between the rendered and ground truth instance masks.}
    \label{fig:method/multi_view_3d_auto_labeling}
\end{figure*}

%% file: related_work/monocular_3d_object_detection/main.tex
\subsection{Monocular 3D Object Detection}
\label{sec:related_work/monocular_3d_object_detction}

Monocular 3D detection is a challenging task due to limited 3D information from monocular imagery. Deep3DBox \cite{DeepBox3D} pioneered this area by regressing relatively stable 3D object properties and combining these estimates with geometric constraints provided by the 2D bounding box. SMOKE \cite{SMOKE} estimates 3D bounding boxes by combining keypoint estimates with regressed 3D box parameters. FCOS3D \cite{FCOS3D} employs a fully convolutional single-stage detector, transforming 7-DoF 3D targets to the image domain and decoupling them as 2D and 3D attributes. To enhance the performance of monocular 3D object detection for truncated objects, MonoFlex \cite{MonoFlex} explicitly decouples truncated objects and adaptively combines multiple approaches for object depth estimation. M3D-RPN \cite{M3D-PRN} and MonoDETR \cite{MonoDETR} also explored the usage of depth cues to improve monocular 3D object detection. The former designed depth-aware convolutional layers that enable location-specific feature extraction and consequently improved 3D scene understanding, while the latter modified the vanilla Transformer \cite{ViT} to be depth-aware to guide the whole detection process by contextual depth cues. Despite these advances, the reliance on expensive and labor-intensive manual annotation on LiDAR point clouds remains a significant limitation.

%% file: related_work/weakly_supervised_3d_object_detection/main.tex
\subsection{Weakly Supervised 3D Object Detection}
\label{sec:related_work/weakly_supervised_3d_object_detection}

Many weakly supervised methods have been proposed to mitigate the high annotation cost in 3D object detection. WS3D \cite{W3D} introduced a LiDAR-based two-stage pipeline where cylindrical object proposals are first generated under weak supervision and then refined using a few labeled object instances. VS3D \cite{VS3D} introduced an unsupervised 3D proposal module that generates object proposals by leveraging normalized point cloud densities. WeakM3D \cite{WeakM3D} introduced a weakly supervised monocular 3D object detection method that leverages the 3D alignment loss between each predicted 3D bounding box and corresponding RoI LiDAR points. Furthermore, it introduced a method to estimate the orientation from RoI LiDAR points based on its statistics. Recently, WeakMono3D \cite{WeakMono3D} leverages projection loss with multi-view and direction consistency, achieving a weakly supervised monocular object detection that relies only on 2D supervision. However, its reliance on 2D direction annotations restricts its applicability to large-scale datasets. Zakharov et al. \cite{Autolabels} proposed an auto-labeling pipeline that integrates an SDF-based differentiable shape renderer and normalized object coordinate spaces (NOCS). However, additional training on synthetic data is still required for shape and coordinate estimation. In contrast, our method purely relies on 2D supervision, eliminating the necessity of synthetic data or 3D supervision.

%% file: related_work/3d_object_detection_with_neural_fields/main.tex
\subsection{3D Object Detection with Neural Fields}
\label{sec:related_work/3d_object_detection_with_neural_fields}

Neural Radiance Fields (NeRF) \cite{NeRF} introduced a new perspective for implicit learning of 3D geometry from posed multi-view images by volume rendering. Building upon the vanilla NeRF, subsequent research has focused on enhancing novel view synthesis \cite{RegNeRF, MipNeRF, MipNeRF360, ZipNeRF} or accelerating volume rendering \cite{InstantNGP, Plenoxels, TensoRF}. NeuS \cite{NeuS} introduced a novel volume rendering scheme to learn a neural SDF representation by introducing a density distribution induced by the SDF. Similarly, VolSDF \cite{VolSDF} defined the volume density function as Laplace’s cumulative distribution function applied to an SDF representation. Many works \cite{NeRFDet, NeRF-RPN, MonoNeRD} recently have attempted to utilize neural fields for 3D object detection. Notably, NeRF-RPN \cite{NeRF-RPN} demonstrated that the 3D bounding boxes of objects in NeRF can be directly regressed without rendering. NeRF-Det \cite{NeRFDet} connects the detection and NeRF branches through a shared MLP, enabling an efficient adaptation of NeRF to detection and yielding geometry-aware volumetric representations. MonoNeRD \cite{MonoNeRD} models scenes with SDFs and renders RGB images and depth maps through volume rendering to obtain intermediate 3D representations for detection. However, these approaches still rely on ground truth 3D labels for supervision. In contrast, we propose the first volume rendering-based weakly supervised 3D object detection framework that relies on multi-view geometry and 2D supervision without any 3D supervision, such as 3D bounding boxes or LiDAR point clouds.

%% file: method/main.tex
\section{Method}
\label{sec:method}
 
\input{method/multi_view_3d_auto_labeling/main}

\input{method/training_of_3d_object_detectors/main}

%% file: method/multi_view_3d_auto_labeling/main.tex
\subsection{Multi-View 3D Auto-Labeling}
\label{sec:method/multi_view_3d_auto_labeling}

The pipeline of our proposed multi-view 3D auto-labeling is illustrated in \cref{fig:method/multi_view_3d_auto_labeling}. In \cref{sec:method/multi_view_3d_auto_labeling/preliminaries}, we review the SDF-based volume rendering formulation, which is the foundation of our method. In \cref{sec:method/multi_view_3d_auto_labeling/problem_definition}, we define the optimization problem for the multi-view 3D auto-labeling. In \cref{sec:method/multi_view_3d_auto_labeling/3d_bounding_box_represented_as_an_sdf}, we introduce the SDF for cuboids, which is used to optimize 3D bounding boxes through rendering. In \cref{sec:method/multi_view_3d_auto_labeling/residual_distance_field}, we introduce a novel neural field named \textit{residual distance field} to fill the spatial gap between the surfaces of each instance and the 3D bounding box. In \cref{sec:method/multi_view_3d_auto_labeling/instance_aware_volumetric_silhouette_rendering}, we introduce instance-aware volumetric silhouette rendering to render the silhouette of each instance based on its SDF. In \cref{sec:method/multi_view_3d_auto_labeling/loss_functions}, we introduce loss functions using instance masks and 2D bounding boxes as weak supervision.

\input{method/multi_view_3d_auto_labeling/preliminaries/main}

\input{method/multi_view_3d_auto_labeling/problem_definition/main}
\input{method/multi_view_3d_auto_labeling/3d_bounding_box_represented_as_an_sdf/main}
\input{method/multi_view_3d_auto_labeling/residual_distance_field/main}
\input{method/multi_view_3d_auto_labeling/instance_aware_volumetric_silhouette_rendering/main}
\input{method/multi_view_3d_auto_labeling/loss_functions/main}

%% file: method/multi_view_3d_auto_labeling/preliminaries/main.tex
\subsubsection{Preliminaries}
\label{sec:method/multi_view_3d_auto_labeling/preliminaries}

\input{method/multi_view_3d_auto_labeling/preliminaries/sdf_based_volumetric_rendering/main}

%% file: method/multi_view_3d_auto_labeling/preliminaries/sdf_based_volumetric_rendering/main.tex
\paragraph{SDF-based Volumetric Rendering}
\label{sec:method/multi_view_3d_auto_labeling/preliminaries/sdf_based_volumetric_rendering}

NeRF \cite{NeRF} represents a 3D scene with neural density and color fields. Given a camera position $\bm{o} \in \mathbb{R}^{3}$ and a ray direction $\bm{d} \in \mathbb{R}^{3}$ emitted from a pixel, the volume rendering scheme integrates the colors of sampled points along the ray as follows:
\begin{align}
    \label{eq:method/multi_view_3d_auto_labeling/preliminaries/sdf_based_volumetric_rendering/integration}
    \hat{\bm{C}}(\bm{o}, \bm{d}) & = \int_{0}^{\infty} w(t) \bm{c}(\bm{r}(t), \bm{d}) \mathrm{d}t \ , \\
    \label{eq:method/multi_view_3d_auto_labeling/preliminaries/sdf_based_volumetric_rendering/volume_density_weight}
    w(t) & = \exp(- \int_{0}^{t} \sigma(\bm{r}(u)) \mathrm{d}u) \sigma(\bm{r}(t)) \ , 
\end{align}
where $\hat{\bm{C}}(\bm{o}, \bm{d}) \in \mathbb{R}^{3}$ denotes the rendered color of the pixel, $\bm{r}(t) = \bm{o} + t\bm{d}$ denotes the ray, $\bm{c}(\bm{p}, \bm{d})$ denotes the color at the position $\bm{p}$ and view direction $\bm{d}$, and $\sigma(\bm{p})$ denotes the \textit{volume density} at the position $\bm{p}$. Since the density field cannot represent the surfaces explicitly, novel SDF-based volume rendering formulation has been introduced in NeuS \cite{NeuS}, where a surface represented by an SDF is reinterpreted as a participating medium represented by a density field, enabling the surface to be rendered through volume rendering. In this formulation, the weight $w(t)$ in \cref{eq:method/multi_view_3d_auto_labeling/preliminaries/sdf_based_volumetric_rendering/volume_density_weight} is re-written by introducing \textit{opaque density} $\rho(t)$ as follows:
\begin{align}
    \label{eq:method/multi_view_3d_auto_labeling/preliminaries/sdf_based_volumetric_rendering/opaque_density_weight}
    w(t) & = \exp(- \int_{0}^{t} \rho(u) \mathrm{d}u) \rho(t) \ , \\
    \label{eq:method/multi_view_3d_auto_labeling/preliminaries/sdf_based_volumetric_rendering/opaque_density}
    \rho(t) & = \max(- \frac{\frac{\mathrm{d} \Phi}{\mathrm{d} t} (\hat{\mathcal{F}}(\bm{r}(t)))}{\Phi(\hat{\mathcal{F}}(\bm{r}(t)))}, 0) \ ,
\end{align}
where $\hat{\mathcal{F}}(\cdot)$ denotes the SDF for the entire scene and $\Phi(\cdot)$ denotes the Sigmoid function. Our proposed instance-aware volumetric silhouette rendering introduced in \cref{sec:method/multi_view_3d_auto_labeling/instance_aware_volumetric_silhouette_rendering} is based on the same SDF-based weight formulation as \cref{eq:method/multi_view_3d_auto_labeling/preliminaries/sdf_based_volumetric_rendering/opaque_density_weight} but integrates instance labels instead of colors along a ray to render instance masks.

%% file: method/multi_view_3d_auto_labeling/problem_definition/main.tex
\subsubsection{Problem Definition}
\label{sec:method/multi_view_3d_auto_labeling/problem_definition}

Given a monocular video consisting of posed frames, each frame annotated with instance masks, our goal is to optimize the 3D bounding box frame by frame without 3D supervision. More specifically, for each \textit{target} frame $t$ in the video, we sample multiple \textit{source} frames $\mathcal{S}$ and optimize the $N$ 3D bounding boxes in the target frame using the instance masks of the source frames as weak supervision, where $N$ denotes the number of instances in the target frame. Please refer to the supplementary material for how to sample $\mathcal{S}$. We parameterize the $n$-th 3D bounding box $\hat{\bm{B}}_{n} \in \mathbb{R}^{8 \times 3}$ in the target frame with a dimension $\hat{\bm{d}}_{n} \in \mathbb{R}^{3}_{+}$, location $\hat{\bm{\ell}}_{n} \in \mathbb{R}^{3}$, and orientation $\hat{\theta}_{n} \in \mathbb{R}$, which is the rotation angle in the bird’s-eye-view. In addition to these parameters for each bounding box, we prepare a learnable instance embedding $\bm{z}_{n} \in \mathbb{R}^{D}$ for each instance and a shared \textit{hypernetwork} parameterized by $\bm{\psi}$ for the \textit{residual distance field} introduced in \cref{sec:method/multi_view_3d_auto_labeling/residual_distance_field}. We stack each parameter group into a single tensor over all the instances, yielding dimensions $\hat{\bm{D}} \in \mathbb{R}^{N \times 3}_{+}$, locations $\hat{\bm{L}} \in \mathbb{R}^{N \times 3}$, orientations $\hat{\bm{\Theta}} \in \mathbb{R}^{N \times 1}$, and instance embeddings $\bm{Z} \in \mathbb{R}^{N \times D}$. Given the loss function $\mathcal{L}$ explained in \cref{sec:method/multi_view_3d_auto_labeling/loss_functions}, we optimize $\hat{\bm{D}}$, $\hat{\bm{L}}$, $\hat{\bm{\Theta}}$, $\bm{Z}$, and $\bm{\psi}$ via stochastic gradient descent as follows:
\begin{align}
    \label{eq:method/multi_view_3d_auto_labeling/problem_definition/optimization_problem}
    ^{*}\!\bm{D}, ^{*}\!\!\bm{L}, ^{*}\!\!\bm{\Theta}, ^{*}\!\!\bm{Z}, ^{*}\!\!\bm{\psi} = \underset{\hat{\bm{D}}, \hat{\bm{L}}, \hat{\bm{\Theta}}, \bm{Z}, \bm{\psi}} {\operatorname{argmin}} \ \mathcal{L}(\hat{\bm{D}}, \hat{\bm{L}}, \hat{\bm{\Theta}}, \bm{Z}, \bm{\psi}) \ .
\end{align}
The optimized 3D bounding boxes $^{*}\!{\bm{B}}$ decoded from $^{*}\!\bm{D}$, $^{*}\!\bm{L}$, and $^{*}\!\bm{\Theta}$ can be used as pseudo labels for training 3D object detectors, as explained in \cref{sec:method/training_of_3d_object_detectors}.

%% file: method/multi_view_3d_auto_labeling/3d_bounding_box_represented_as_an_sdf/main.tex
\subsubsection{3D Bounding Box Represented as an SDF}
\label{sec:method/multi_view_3d_auto_labeling/3d_bounding_box_represented_as_an_sdf}

To optimize the 3D bounding boxes in each target frame by solving \cref{eq:method/multi_view_3d_auto_labeling/problem_definition/optimization_problem} through rendering, we represent the surface of each 3D bounding box as a signed distance field (SDF), which is one of the most common surface representations. An SDF is represented as a function $\mathcal{F} : \mathbb{R}^{3} \rightarrow \mathbb{R}$ that maps a spatial position $\bm{p} \in \mathbb{R}^{3}$ to its signed distance to the closest point on the surface, indicating that the zero-level set $\{\bm{p} \in \mathbb{R}^{3} \mid \mathcal{F}(\bm{p}) = 0\}$ represents the surface itself. The SDF for a cuboid parameterized by a dimension $\bm{d} \in \mathbb{R}^{3}_{+}$, location $\bm{\ell} \in \mathbb{R}^{3}$, and orientation $\bm{R} \in \text{SO}(3)$ can be derived theoretically and is denoted by $\mathcal{B}(\cdot ; \bm{d}, \bm{\ell}, \bm{R})$. Please refer to the supplementary material for how to derive this formula. 
 

%% file: method/multi_view_3d_auto_labeling/residual_distance_field/main.tex
\subsubsection{Residual Distance Field}
\label{sec:method/multi_view_3d_auto_labeling/residual_distance_field}

In general, the shape of each instance is not a cuboid. Therefore, if we leverage only the cuboid SDF introduced in \cref{sec:method/multi_view_3d_auto_labeling/3d_bounding_box_represented_as_an_sdf} to render the surface of each instance, it is no longer possible to render an accurate silhouette due to the spatial gap between the surfaces of the instance and the 3D bounding box, leading to unreliable feedback signals during optimization. Therefore, we propose a novel neural field named \textit{residual distance field} (RDF), which models the residual between the signed distances to the surfaces of the instance and the 3D bounding box. Let $\mathcal{F}_{n}(\cdot)$ be the \textit{true} SDF of the surface bounded by the $n$-th 3D bounding box $\hat{\bm{B}}_{n}$ whose SDF is given by $\hat{\mathcal{B}}_{n}(\cdot) = \mathcal{B}(\cdot ; \hat{\bm{d}}_{n}, \hat{\bm{\ell}}_{n}, \bm{R}_{y}(\hat{\theta}_{n}))$, where $\bm{R}_{y}(\theta)$ denotes the rotation matrix around the $y$-axis by an angle $\theta$. For any point $\bm{p} \in \mathbb{R}^{3}$, we define the RDF as $\hat{\mathcal{R}}_{n}(\bm{p}) \coloneqq \mathcal{F}_{n}(\bm{p}) - \hat{\mathcal{B}}_{n}(\bm{p})$. Here, based on the definition of a 3D bounding box that it encloses the corresponding instance, $\forall \bm{p} \in \mathbb{R}^{3} : \hat{\mathcal{R}}_{n}(\bm{p}) \ge 0 $ is required. One straightforward way to model the RDF for each instance is to train $N$ individual networks. However, as objects belonging to the same semantic class often have similar geometric shapes, we train a single \textit{hypernetwork} \cite{Hypernetworks} that regresses the weights of the neural RDF directly from an instance embedding instead. Given the $n$-th instance embedding $\bm{z}_{n} \in \mathbb{R}^{D}$, where $D$ denotes the number of dimensions, the $n$-th neural RDF is given by:
\begin{align}
    \label{eq:method/multi_view_3d_auto_labeling/residual_distance_field/hypernetwork}
    \hat{\mathcal{R}}_{n}(\bm{p}) = \sigma(\mathcal{G}(\bm{p} ;  \bm{\phi}_{n})) \ , \\
    \bm{\phi}_{n} = \mathcal{H}(\bm{z}_{n} ; \bm{\psi}) \ , 
\end{align}
where $\mathcal{G}(\cdot ; \bm{\phi}_{n})$ denotes the $n$-th neural network parameterized by $\bm{\phi}_{n}$, $\mathcal{H}(\cdot ; \bm{\psi})$ denotes the shared hypernetwork parameterized by $\bm{\psi}$, and $\sigma(\cdot)$ denotes the Softplus function to force the residual distance to be positive based on the definition of a bounding box. For the instance-aware volumetric silhouette rendering introduced in \cref{sec:method/multi_view_3d_auto_labeling/instance_aware_volumetric_silhouette_rendering}, we employ $\hat{\mathcal{F}}_{n}(\bm{p}) = \hat{\mathcal{B}}_{n}(\bm{p}) + \hat{\mathcal{R}}_{n}(\bm{p})$ as the SDF of each instance that represents an arbitrary surface bounded by the 3D bounding box $\hat{\bm{B}}_{n}$ for more accurate silhouette rendering. 

%% file: method/multi_view_3d_auto_labeling/instance_aware_volumetric_silhouette_rendering/main.tex
\input{method/multi_view_3d_auto_labeling/instance_aware_volumetric_silhouette_rendering/figure}

\subsubsection{\scalebox{0.98}{Instance-Aware Volumetric Silhouette Rendering}}
\label{sec:method/multi_view_3d_auto_labeling/instance_aware_volumetric_silhouette_rendering}

This section details our approach to optimizing 3D bounding boxes through volumetric rendering. The core idea is to render instance masks and compare them with ground truth instance masks. To achieve this, we propose a novel SDF-based \textit{instance-aware} volumetric silhouette rendering, where instance labels instead of colors are integrated along a ray based on the same SDF-based volume rendering formulation as \cref{eq:method/multi_view_3d_auto_labeling/preliminaries/sdf_based_volumetric_rendering/opaque_density_weight} as follows:
\begin{align}
    \label{eq:method/multi_view_3d_auto_labeling/instance_aware_volumetric_silhouette_rendering/integration}
    \hat{\bm{S}}(\bm{o}, \bm{d}) & = \int_{0}^{\infty} w(t) \bm{s}(\bm{r}(t)) dt \ , \\
    \label{eq:method/multi_view_3d_auto_labeling/instance_aware_volumetric_silhouette_rendering/average_instance_label}
    \bm{s}(\bm{p}) & = \sum_{n=1}^{N} \text{softmin}([\hat{\mathcal{F}}_{n}(\bm{p})]_{n=1}^{N})_{n} \cdot \bm{y}_{n} \ , 
\end{align}
where, $\hat{\bm{S}}(\bm{o}, \bm{d}) \in [0, 1]^{N}$ denotes the rendered soft instance label, $\bm{s}(\bm{p}) \in [0, 1]^{N}$ denotes the weighted average instance label at the position $\bm{p}$ indicating how relatively close the position $\bm{p}$ is to each instance, and $\bm{y}_{n} \in \{0, 1\}^{N}$ denotes the one-hot instance label of the $n$-th instance. $\hat{\mathcal{F}}_{n}(\cdot)$ denotes the SDF of the $n$-th instance introduced in \cref{sec:method/multi_view_3d_auto_labeling/residual_distance_field}. $w(t)$ in \cref{eq:method/multi_view_3d_auto_labeling/instance_aware_volumetric_silhouette_rendering/integration} is derived from \cref{eq:method/multi_view_3d_auto_labeling/preliminaries/sdf_based_volumetric_rendering/opaque_density_weight}. To compute \cref{eq:method/multi_view_3d_auto_labeling/preliminaries/sdf_based_volumetric_rendering/opaque_density_weight}, we model the entire scene as the union of all the object surfaces, i.e., $\hat{\mathcal{F}}(\bm{p}) = \min(\hat{\mathcal{F}}_{1}(\bm{p}), \ldots, \hat{\mathcal{F}}_{N}(\bm{p}))$. This mechanism enables us to render instance masks considering the geometric relationships among instances, such as occlusions.

%% file: method/multi_view_3d_auto_labeling/instance_aware_volumetric_silhouette_rendering/figure.tex
\begin{figure}[t]
    \centering
    \includegraphics[width=\linewidth]{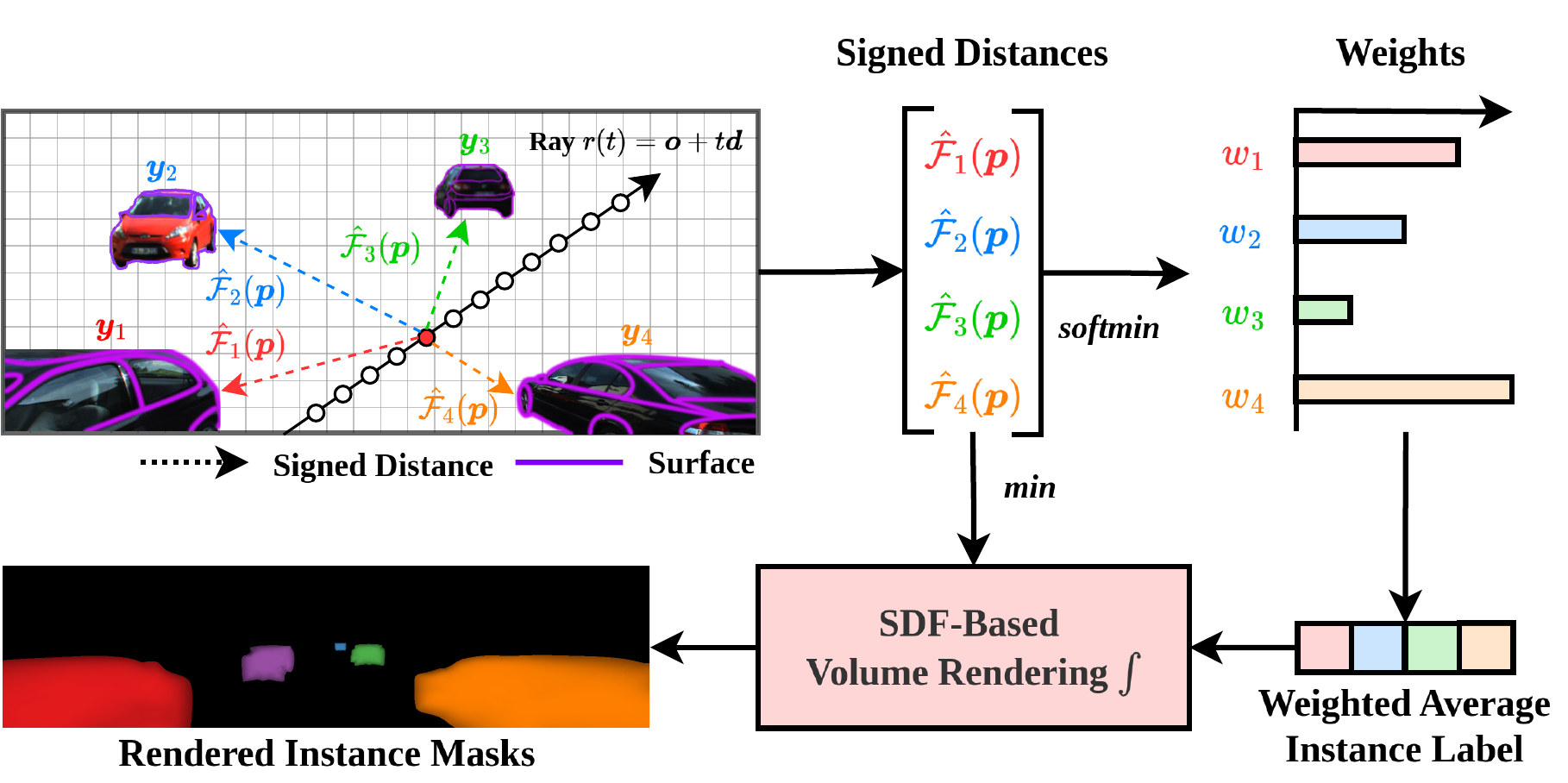}
    \caption{Illustration of our proposed instance-aware volumetric silhouette rendering. The instance labels are averaged for each sampled point along a ray based on the signed distance to each instance. The averaged instance labels are integrated along the ray based on the SDF-based volume rendering formulation \cite{NeuS}.}
    \label{fig:method/multi_view_3d_auto_labeling/instance_aware_volumetric_silhouette_rendering}
\end{figure}

%% file: method/multi_view_3d_auto_labeling/loss_functions/main.tex
\subsubsection{Loss Functions}
\label{sec:method/multi_view_3d_auto_labeling/loss_functions}

We optimize the parameters of the 3D bounding boxes $\hat{\bm{\Omega}} = \{\hat{\bm{D}}, \hat{\bm{L}}, \hat{\bm{\Theta}}\}$ in the target frame along with the instance embeddings $\bm{Z}$ and parameter $\bm{\psi}$ of the hypernetwork $\mathcal{H}(\cdot ; \bm{\psi})$. The final loss $\mathcal{L}$ is defined as the combination of the \textit{multi-view projection loss} $\mathcal{L}_{\text{proj}}$, \textit{multi-view silhouette loss} $\mathcal{L}_{\text{slh}}$, and Eikonal regularization $\mathcal{L}_{\text{reg}}$ \cite{NeuS} as follows:
\begin{align}
    \label{sec:method/multi_view_3d_auto_labeling/loss_functions/total_loss}
    \mathcal{L}(\hat{\bm{\Omega}}, \bm{Z}, \bm{\psi}) 
    & = \lambda_{\text{proj}} \mathcal{L}_{\text{proj}}(\hat{\bm{\Omega}}) + \lambda_{\text{slh}} \mathcal{L}_{\text{slh}}(\hat{\bm{\Omega}}, \bm{Z}, \bm{\psi}) \nonumber  \\ 
    & + \lambda_{\text{reg}} \mathcal{L}_{\text{reg}}(\hat{\bm{\Omega}}, \bm{Z}, \bm{\psi}) \ ,  
\end{align}
where $\lambda_{\text{proj}}$, $\lambda_{\text{slh}}$, and $\lambda_{\text{reg}}$ denote the loss weights. However, although the multi-view projection loss $\mathcal{L}_{\text{proj}}$ and multi-view silhouette loss $\mathcal{L}_{\text{slh}}$ are based on 2D supervision, the 3D-2D correspondence to define the losses is not obvious. Therefore, we find an optimal bipartite matching between the optimized 3D bounding boxes and ground truth 2D bounding boxes by the Hungarian algorithm \cite{Hungarian}. Each pair-wise matching cost is defined as the projection loss introduced in this section, but it is computed for not all the source frames but only the target frame. We assume that the ground truth 2D bounding boxes and instance masks have already been reordered based on the optimal permutation for simplicity.

\input{method/multi_view_3d_auto_labeling/loss_functions/multi_view_projection_loss/main}
\input{method/multi_view_3d_auto_labeling/loss_functions/multi_view_silhouette_loss/main}

%% file: method/multi_view_3d_auto_labeling/loss_functions/multi_view_projection_loss/main.tex
\paragraph{Multi-View Projection Loss}
\label{sec:method/multi_view_3d_auto_labeling/loss_functions/multi_view_projection_loss}

Although introducing the residual distance field enables us to model an arbitrary surface bounded by a 3D bounding box, it introduces another problem: the 3D bounding box can grow indefinitely without any constraint. To address this problem, we employ the ground truth 2D bounding box as a constraint to keep the 3D bounding box tight. The multi-view projection loss $\mathcal{L}_{\text{proj}}$ is defined as the average distance between the \textit{projected} and ground truth 2D bounding boxes as follows:
\begin{align}
    \label{eq:method/multi_view_3d_auto_labeling/loss_functions/multi_view_projection_loss/multi_view_projection_loss}
    \mathcal{L}_{\text{proj}}(\hat{\bm{\Omega}}) 
    & = \alpha \sum_{i \in \mathcal{S}} \sum_{n=1}^{N} \|\hat{\bm{B}}^{\text{2D}}_{in} - \bm{B}^{\text{2D}}_{in}\|_{\rm{H}} \nonumber \\ 
    & - \beta \sum_{i \in \mathcal{S}} \sum_{n=1}^{N} \text{DIoU}(\hat{\bm{B}}^{\text{2D}}_{in}, \bm{B}^{\text{2D}}_{in}) \ , 
\end{align}
where $\|\cdot\|_{\rm{H}}$ denotes the Huber loss, $\text{DIoU}(\cdot, \cdot)$ denotes the Distance-IoU \cite{DIoU}, and $\alpha, \beta$ denote balancing coefficients. The projected 2D bounding box $\hat{\bm{B}}^{\text{2D}}_{in}$ is defined as the rectangle with minimal area enclosing the projected vertices $\hat{\bm{V}}^{\text{2D}}_{in} \propto \hat{\bm{B}}_{n} \bm{E}_{i}^{T} \bm{K}_{i}^{T}$, where $\bm{E}_{i}$ and $\bm{K}_{i}$ denote the extrinsic and intrinsic matrices for frame $i$, respectively. 

%% file: method/multi_view_3d_auto_labeling/loss_functions/multi_view_silhouette_loss/main.tex
\paragraph{Multi-View Silhouette Loss} 
\label{sec:method/multi_view_3d_auto_labeling/loss_functions/multi_view_silhouette_loss}

The multi-view silhouette loss is defined as the cross entropy between the rendered and ground truth instance masks. Since the spatial gap between the surfaces of each instance and the 3D bounding box is modeled by the residual distance field introduced in \cref{sec:method/multi_view_3d_auto_labeling/residual_distance_field}, the 3D bounding box constrained to tightly fit the ground truth 2D bounding box by the multi-view projection loss $\mathcal{L}_{\text{proj}}$ is further refined by the multi-view silhouette loss $\mathcal{L}_{\text{slh}}$, which is given by:
\begin{align}
    \label{eq:method/multi_view_3d_auto_labeling/loss_functions/multi_view_silhouette_loss/multi_view_silhouette_loss}
    \mathcal{L}_{\text{slh}}(\hat{\bm{\Omega}}, \bm{Z}, \bm{\psi}) = \sum_{i \in \mathcal{S}} \sum_{j=1}^{R_{i}} \text{CE}(\hat{\bm{S}}(\bm{o}_{i}, \bm{d}_{ij}), \bm{S}_{ij}) \ , 
\end{align}
where $\bm{o}_{i} \in \mathbb{R}^{3}$, $\bm{d}_{ij} \in \mathbb{R}^{3}$, and $\bm{S}_{ij} \in \{0, 1\}^{N}$ denote the camera position, ray direction, and ground truth instance label at the $j$-th sampled pixel in frame $i$, respectively. $\hat{\bm{S}}(\bm{o}_{i}, \bm{d}_{ij})$ denotes the rendered instance label based on \cref{eq:method/multi_view_3d_auto_labeling/instance_aware_volumetric_silhouette_rendering/integration}. $\text{CE}(\cdot, \cdot)$ denotes the cross entropy loss. $R_{i}$ denotes the number of rays sampled for frame $i$. Please refer to the supplementary material for how to determine $R_{i}$.

%% file: method/training_of_3d_object_detectors/main.tex
\input{method/training_of_3d_object_detectors/confidence_assignment/figure}

\subsection{Training of 3D Object Detectors}
\label{sec:method/training_of_3d_object_detectors}

Once the 3D bounding boxes are optimized by the proposed multi-view 3D auto-labeling, they can serve as pseudo labels for training 3D object
detectors. To incorporate the quality of each pseudo label into the training of 3D object detectors, we introduce a simple but effective confidence assignment method in \cref{sec:method/training_of_3d_object_detectors/confidence_assignment} and confidence-based weighted loss for bounding box regression in \cref{sec:method/training_of_3d_object_detectors/confidence_based_weighted_loss}.

\input{method/training_of_3d_object_detectors/confidence_assignment/main}
\input{method/training_of_3d_object_detectors/confidence_based_weighted_loss/main}

%% file: method/training_of_3d_object_detectors/confidence_assignment/figure.tex
\begin{figure}[t]
    \setlength{\abovecaptionskip}{1mm}
    \begin{minipage}[b]{0.49\columnwidth}
        \centering
        \includegraphics[width=1.0\columnwidth]{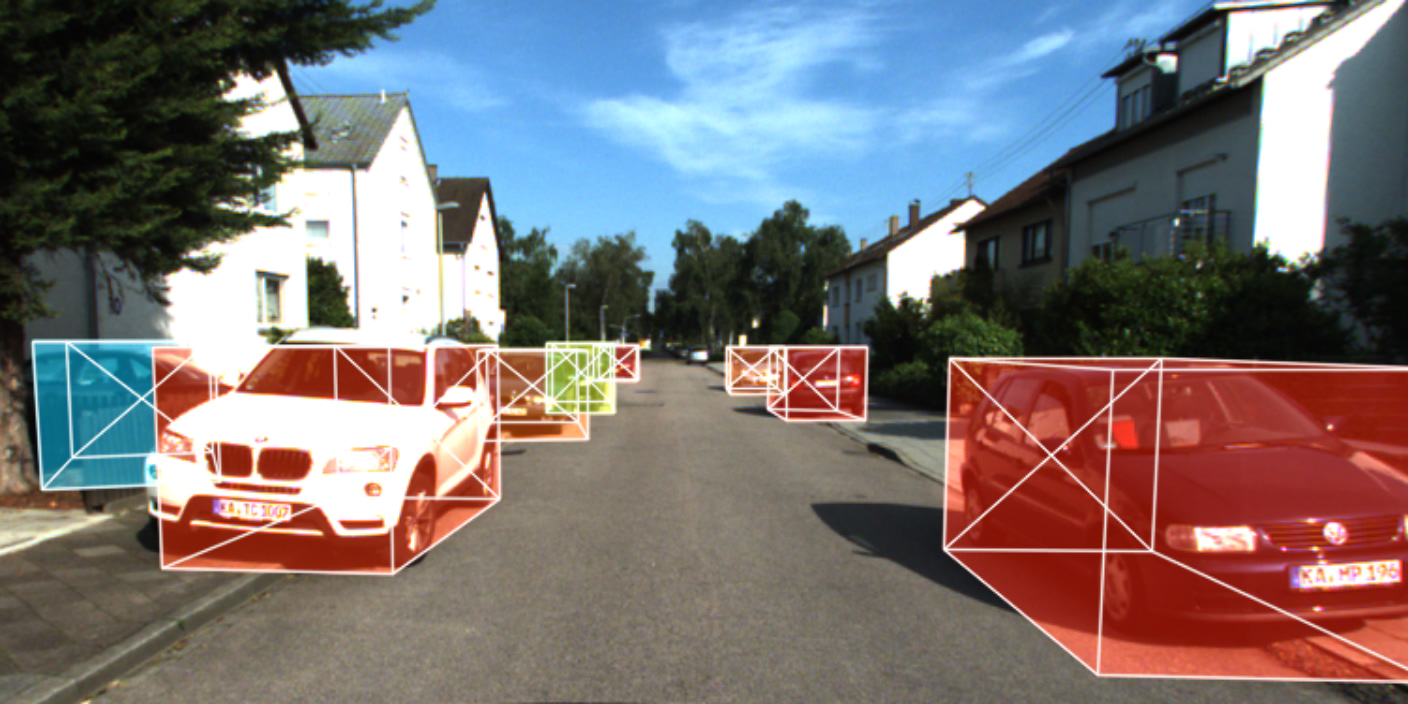}
        \subcaption{Static Scene}
    \end{minipage}
    \begin{minipage}[b]{0.49\columnwidth}
        \centering
        \includegraphics[width=1.0\columnwidth]{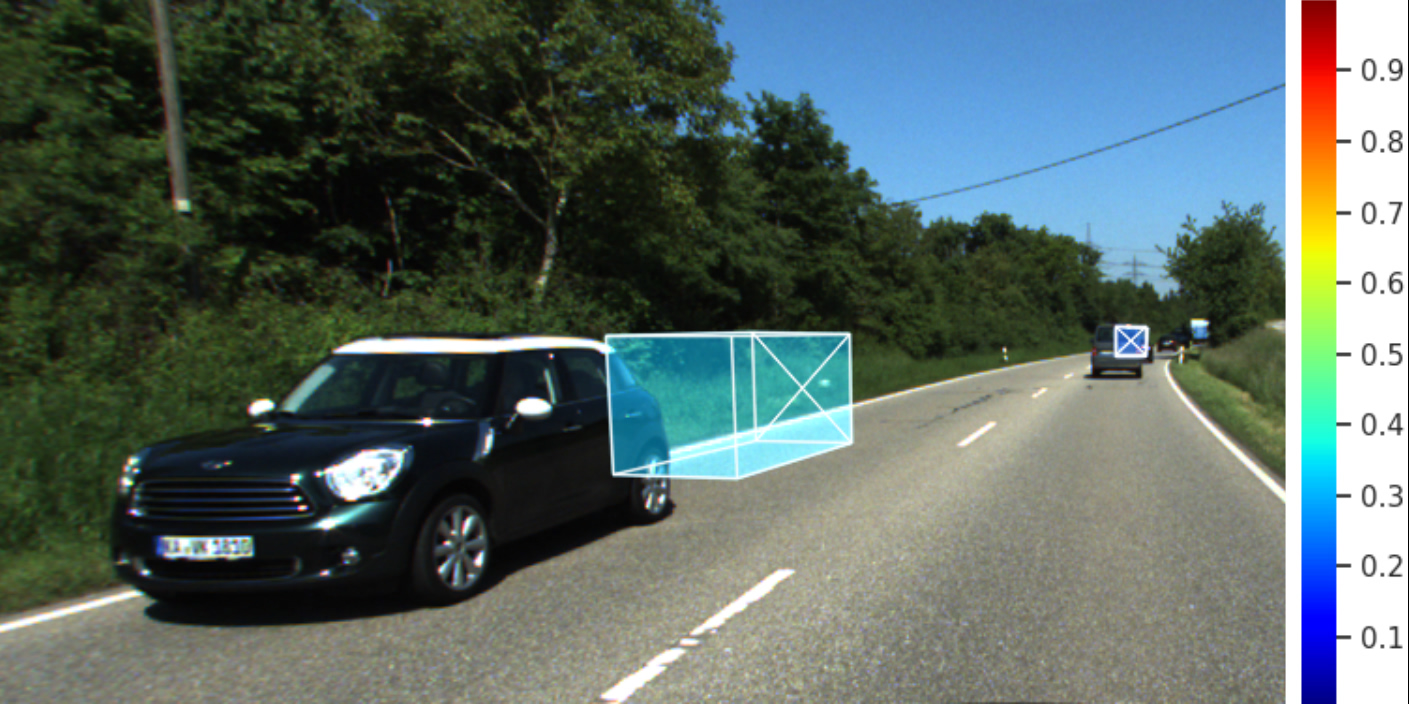}
        \subcaption{Dynamic Scene}
    \end{minipage}
    \caption{Comparison of the confidence scores in static and dynamic scenes. It can be seen that the confidence scores are lower for dynamic, occluded, or truncated objects, indicating less influence on the subsequent training of 3D object detectors.}
    \label{fig:method/training_of_3d_object_detectors/confidence_assignment}
\end{figure}

%% file: method/training_of_3d_object_detectors/confidence_assignment/main.tex
\subsubsection{Confidence Assignment}
\label{sec:method/training_of_3d_object_detectors/confidence_assignment}

As shown in \cref{fig:method/training_of_3d_object_detectors/confidence_assignment}, the 3D bounding boxes optimized by the proposed auto-labeling are not reliable for dynamic, occluded, or truncated objects. Therefore, we propose a confidence assignment method based on the multi-view projection loss. First, for each target frame $t$, we identify a set of source frames $\mathcal{I}$ such that all the instances in the target frame are visible from every frame in the set. Next, we find an optimal bipartite matching between the optimized 3D bounding boxes and ground truth 2D bounding boxes by the Hungarian algorithm. The cost matrix $\bm{Q} \in \mathbb{R}^{N \times N}$ is defined as the pairwise IoUs between the projected and ground truth 2D bounding boxes averaged over all the source frames, as follows: 
\begin{align}
    \label{eq:method/training_of_3d_object_detectors/confidence_assignment/cost_matrix}
    \bm{Q}_{nm} =  \frac{1}{|\mathcal{I}|} \sum_{i \in \mathcal{I}} 1 - \text{IoU}(^{*}\!\bm{B}^{\text{2D}}_{in}, \bm{B}^{\text{2D}}_{im}) \ ,
\end{align}
where $^{*}\!\bm{B}^{\text{2D}}_{in}$ denotes the $n$-th projected 2D bounding box in frame $i$, which is obtained by projecting the $n$-th optimized 3D bounding box $^{*}\!\bm{B}_{n}$ onto frame $i$, and $\bm{B}^{\text{2D}}_{im}$ denotes the $m$-th ground truth 2D bounding box in frame $i$. Once an optimal permutation matrix $\bm{P} \in \{0, 1\}^{N \times N}$ is obtained, the confidence scores $^{*}\!\bm{C} \in [0, 1]^{N}$ for the optimized 3D bounding boxes $^{*}\!\bm{B}$ are given as follows:
\begin{align}
    \label{eq:method/training_of_3d_object_detectors/confidence_assignment/confidence_scores}
    ^{*}\!\bm{C} = \frac{1}{|\mathcal{I}|} \sum_{i \in \mathcal{I}} \text{IoU}(^{*}\!\bm{B}^{\text{2D}}_{i}, \bm{P} \bm{B}^{\text{2D}}_{i}) \ .
\end{align}

%% file: method/training_of_3d_object_detectors/confidence_based_weighted_loss/main.tex
\subsubsection{Confidence-based Weighted Loss}
\label{sec:method/training_of_3d_object_detectors/confidence_based_weighted_loss}

The 3D bounding boxes optimized by the proposed multi-view 3D auto-labeling serve as pseudo labels for most 3D object detectors, which leverage 3D bounding boxes and semantic class labels as supervision, without any modification of the architectures, loss functions, and training procedures, except for the confidence-based loss weighting. We incorporate the confidence scores computed by \cref{eq:method/training_of_3d_object_detectors/confidence_assignment/confidence_scores} into only the regression loss. Given a regression loss function $\mathcal{L}_{\text{box}}$, its confidence-based weighted version $\tilde{\mathcal{L}}_{\text{box}}$ is given by:
\begin{align}
    \label{sec:method/training_of_3d_object_detectors/confidence_based_weighted_loss/confidence_based_weighted_loss}
    \tilde{\mathcal{L}}_{\text{box}}(\hat{\bm{B}}, ^{*}\!\!\bm{B}, ^{*}\!\!\bm{C}) = \sum_{m=1}^{M} \ ^{*}\!\bm{C}_{\pi(m)} \mathcal{L}_{\text{box}}(\hat{\bm{B}}_{m}, ^{*}\!\!\bm{B}_{\pi(m)}) ,
\end{align}
where $\hat{\bm{B}}$ denotes the predicted 3D bounding boxes, $^{*}\!\bm{B}$ denotes the 3D bounding boxes optimized by the proposed auto-labeling, and $^{*}\!\bm{C}$ denotes the corresponding confidence scores. $M$ denotes the number of positive anchors, and $\pi(\cdot)$ denotes the label assigner that maps the index of an anchor to that of the matched ground truth.

%% file: experiments/main.tex
\section{Experiments}
\label{sec:experiments}

\input{experiments/dataset/main}
\input{experiments/implementation_details/main}

\input{experiments/ablation_study/main}

\input{experiments/evaluation_results/main}

%% file: experiments/dataset/main.tex
\subsection{Dataset}
\label{sec:experiments/datasets}

We use the KITTI-360 \cite{KITTI-360} dataset for our experiments, splitting it into training (43,855 images), validation (1,173 images), and test sets (2,531 images). We follow the same evaluation protocol as the KITTI dataset \cite{KITTI}. However, as occlusion and truncation labels are not available for the KITTI-360 dataset, unlike the KITTI dataset, we consider only two difficulty levels, namely \textit{Easy} and \textit{Hard}, based on whether the height of each ground truth 2D bounding box is greater than $40$ and $25$ in pixels, respectively. Following prior works, we evaluate our method on only category \textit{Car}.

%% file: experiments/implementation_details/main.tex
\subsection{Implementation Details}
\label{sec:experiments/implementation_details}

\input{experiments/implementation_details/multi_view_3d_auto_labeling/main}
\input{experiments/implementation_details/monocular_3d_object_detection/main}

%% file: experiments/implementation_details/multi_view_3d_auto_labeling/main.tex
\subsubsection{Multi-View 3D Auto-Labeling}
\label{sec:experiments/implementation_details/multi_view_3d_auto_labeling}

We sample $16$ source frames for each target frame. We sample $1000$ rays across all the source frames, i.e., $\sum_{i \in \mathcal{S}} R_{i} = 1000$, at each iteration based on the ground truth instance masks. Please refer to the supplementary material for more details. We employ the same hierarchical volume sampling as NeRF \cite{NeRF} and sample $100$ query points for both \textit{coarse} and \textit{fine} sampling. Unlike NeRF, both coarse and fine samples are drawn from the single scene SDF rather than two distinct ones. We set the number of dimensions of each instance embedding as $D = 256$. The neural RDF $\mathcal{G}$ and hypernetwork $\mathcal{H}$ are implemented as MLPs with four hidden layers, each of which has $256$ and $16$ channels, respectively. We use the Adam optimizer \cite{adam} and the learning rates are decayed exponentially from $1e^{-2}$, $1e^{-3}$, and $1e^{-4}$ to $1e^{-4}$, $1e^{-5}$, and $1e^{-6}$ for the box parameters $\hat{\bm{\Omega}}$, instance embeddings $\bm{Z}$, and parameter $\bm{\psi}$ of the hypernetwork $\mathcal{H}(\cdot ; \bm{\psi})$ over $3000$ iterations, respectively. For the loss weights, we set $\alpha = 1.0$, $\beta = 0.1$, $\lambda_{\text{proj}} = 1.0$, $\lambda_{\text{slh}} = 1.0$, and $\lambda_{\text{reg}} = 0.01$.

%% file: experiments/implementation_details/monocular_3d_object_detection/main.tex
\subsubsection{Monocular 3D Object Detection}
\label{sec:experiments/implementation_details/monocular_3d_object_detection}

To compare our method and Autolabels \cite{Autolabels} with WeakM3D \cite{WeakM3D}, we modify the architecture of WeakM3D so that it can be trained in a supervised manner using pseudo labels. More specifically, we train dimension and confidence heads in addition to the existing location and orientation heads using the same supervised loss as MonoDIS \cite{MonoDIS}. We call this model S-WeakM3D. For both WeakM3D and S-WeakM3D, ground truth 2D bounding boxes and instance masks are used for RoIAlign \cite{MaskRCNN} and generating LiDAR points on each object surface during training, respectively. During inference, we employ Cascade Mask R-CNN \cite{CascadeMaskRCNN} with InternImage-XL \cite{InternImage} as the off-the-shelf 2D detector to provide 2D bounding boxes for RoIAlign. Please refer to the supplementary material for more details.

%% file: experiments/ablation_study/main.tex
\subsection{Ablation Study}
\label{sec:experiments/ablation_study}

\input{experiments/ablation_study/multi_view_3d_auto_labeling/table}
\input{experiments/ablation_study/confidence_assignment/table}
\input{experiments/ablation_study/multi_view_3d_auto_labeling/main}
\input{experiments/ablation_study/confidence_assignment/main}

%% file: experiments/ablation_study/multi_view_3d_auto_labeling/table.tex
\begin{table}[t]
    \caption{Ablation study on the KITTI-360 training set to verify the effectiveness of each component in our proposed multi-view 3D auto-labeling. Just one randomly selected sequence is used.}
    \label{tab:experiments/ablation_study/multi_view_3d_auto_labeling}
    \centering
    \setlength{\tabcolsep}{0.8mm}
    \scalebox{0.85}{
        \begin{tabular}{ccccccc}
            \toprule
            \multicolumn{3}{c}{Components} & \multicolumn{2}{c}{AP$_{\text{BEV}}$/AP$_{\text{3D}}$@0.3} & \multicolumn{2}{c}{AP$_{\text{BEV}}$/AP$_{\text{3D}}@0.5$@0.5} \\
            \cmidrule(lr){1-3} \cmidrule(lr){4-5} \cmidrule(lr){6-7}
            $\mathcal{L}_{\text{proj}}$ & $\mathcal{L}_{\text{slh}}$ & RDF & Easy & Hard & Easy & Hard \\
            \midrule
            \cmark &        &        & 60.77/54.88 & 63.99/57.66 & 37.38/23.33 & 37.44/24.82 \\
            \cmark & \cmark &        & 63.84/56.11 & 60.86/56.87 & 41.73/26.84 & 39.22/25.58 \\
            \cmark & \cmark & \cmark & \textbf{73.84}/\textbf{66.64} & \textbf{73.22}/\textbf{66.32} & \textbf{46.35}/\textbf{31.11} & \textbf{43.07}/\textbf{30.16} \\
            \bottomrule
        \end{tabular}
    }
    \vspace{-0mm}
\end{table}

%% file: experiments/ablation_study/confidence_assignment/table.tex
\begin{table}[t]
    \caption{Ablation study on the KITTI-360 test set to verify the effectiveness of the confidence scores against monocular 3D object detection. S-WeakM3D is used as a monocular 3D object detector.}
    \label{tab:experiments/ablation_study/confidence_assignment}
    \centering
    \scalebox{0.85}{
        \begin{tabular}{ccccc}
            \toprule
            & \multicolumn{2}{c}{AP$_{\text{BEV}}$/AP$_{\text{3D}}$@0.3} & \multicolumn{2}{c}{AP$_{\text{BEV}}$/AP$_{\text{3D}}$@0.5} \\
            \cmidrule(lr){2-3} \cmidrule(lr){4-5}
            Conf. & Easy  & Hard  & Easy  & Hard  \\
            \midrule
                   & 38.67/31.62 & 31.25/23.88 & 10.95/8.25 & 5.37/4.41 \\
            \cmark & \textbf{51.09}/\textbf{42.94} & \textbf{41.27}/\textbf{33.97} & \textbf{19.50}/\textbf{11.91} & \textbf{14.39}/\textbf{8.46} \\
            \bottomrule
        \end{tabular}
    }
    \vspace{-2mm}
\end{table}

%% file: experiments/ablation_study/multi_view_3d_auto_labeling/main.tex
\subsubsection{Multi-View 3D Auto-Labeling}
\label{sec:experiments/ablation_study/multi_view_3d_auto_labeling}

We conduct an ablation study to demonstrate the effectiveness of each component in our proposed multi-view 3D auto-labeling. As can be seen from \cref{tab:experiments/ablation_study/multi_view_3d_auto_labeling}, the multi-view projection loss $\mathcal{L}_{\text{proj}}$ serves as a practical baseline on its own. The multi-view silhouette loss $\mathcal{L}_{\text{slh}}$ further boosts the quality of the pseudo labels a little bit, while the spatial gap between the surfaces of each instance and the 3D bounding box limits further improvements. However, the residual distance field (RDF) can significantly boost the quality of the pseudo labels by addressing this problem. The systematic improvements with the inclusion of each component indicate their effectiveness, resulting in a precise auto-labeling system for weakly supervised 3D object detection.

%% file: experiments/ablation_study/confidence_assignment/main.tex
\subsubsection{Confidence Assignment}
\label{sec:experiments/ablation_study/confidence_assignment}

Due to the unreliability of the 3D bounding boxes optimized by the proposed auto-labeling for dynamic, occluded, or truncated objects, we conduct an ablation study to demonstrate the effectiveness of our proposed confidence assignment. \cref{tab:experiments/ablation_study/confidence_assignment} highlights substantial enhancement in detection performance when using the confidence-incorporated pseudo labels compared with the baseline, demonstrating the effectiveness of the proposed confidence assignment.

%% file: experiments/evaluation_results/main.tex
\input{experiments/evaluation_results/multi_view_3d_auto_labeling/figure}
\input{experiments/evaluation_results/multi_view_3d_auto_labeling/table}
\input{experiments/evaluation_results/monocular_3d_object_detection/weakly_supervised_setting/table}
\input{experiments/evaluation_results/monocular_3d_object_detection/semi_supervised_setting/table}

\subsection{Evaluation Results}
\label{sec:experiments/evaluation_results}

\input{experiments/evaluation_results/multi_view_3d_auto_labeling/main}
\input{experiments/evaluation_results/monocular_3d_object_detection/main}

%% file: experiments/evaluation_results/multi_view_3d_auto_labeling/figure.tex
\begin{figure}[!t]
  \begin{minipage}[b]{1.0\linewidth}
    \centering
    \includegraphics[width=1.0\linewidth]{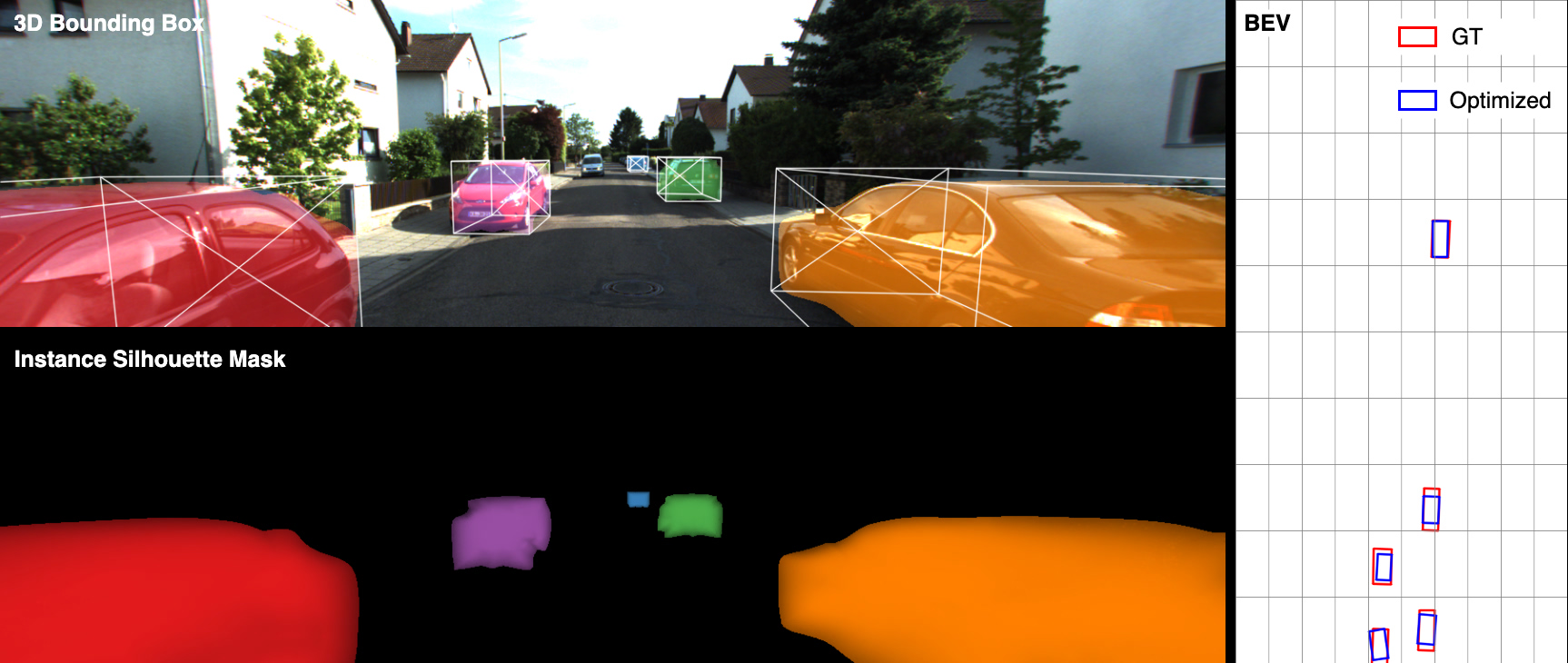}
  \end{minipage}
  \caption{Visualization results of the optimized 3D bounding boxes (1st row) and rendered instance masks (2nd row). We assign a unique color to each instance, and each pixel is colored as the weighted summation based on the rendered soft instance label.}
  \label{fig:experiments/evaluation_results/multi_view_3d_auto_labeling}
  \vspace{-0mm}
\end{figure}

%% file: experiments/evaluation_results/multi_view_3d_auto_labeling/table.tex
\begin{table}[t]
    \caption{Evaluation results of our proposed multi-view 3D auto-labeling on the KITTI-360 training set compared with the LiDAR-based monocular 3D auto-labeling proposed in Autolabels \cite{Autolabels}. $^{*}$Reproduced with the official code.}
    \label{tab:experiments/evaluation_results/multi_view_3d_auto_labeling}
    \centering
    \renewcommand{\footnoterule}{\empty}
    \renewcommand{\thefootnote}{\fnsymbol{footnote}}
    \renewcommand{\thempfootnote}{\fnsymbol{mpfootnote}}
    \setlength{\tabcolsep}{0.8mm}
    \scalebox{0.8}{
        \begin{tabular}{cccccc}
            \toprule
            & & \multicolumn{2}{c}{AP$_{\text{BEV}}$/AP$_{\text{3D}}@$@0.3} & \multicolumn{2}{c}{AP$_{\text{BEV}}$/AP$_{\text{3D}}$@0.5} \\
            \cmidrule(lr){3-4} \cmidrule(lr){5-6}
            Conf. & Method                                       & Easy  & Hard  & Easy  & Hard  \\
            \midrule
            \multirow{2}{*}{$\ge 0.0$} & Autolabels\footnotemark[1] & 71.24/15.09 & 67.33/11.42 & \textbf{51.85}/4.65 & \textbf{46.10}/2.92 \\
            & VSRD                                                             & \textbf{75.03}/\textbf{68.53} & \textbf{72.11}/\textbf{65.64} & 47.12/\textbf{35.25} & 43.91/\textbf{32.64} \\
            \midrule
            \multirow{2}{*}{$\ge 0.8$} & Autolabels\footnotemark[1] & 75.56/17.18 & 72.23/12.19 & \textbf{58.64}/5.77 & 52.99/3.75 \\
            & VSRD                                   & \textbf{84.54}/\textbf{80.25} & \textbf{81.66}/\textbf{77.37} & 58.57/\textbf{45.76} & \textbf{55.09}/\textbf{44.17} \\
            \bottomrule
        \end{tabular}
    }
    \vspace{-2mm}
\end{table}

%% file: experiments/evaluation_results/monocular_3d_object_detection/weakly_supervised_setting/table.tex
\begin{table*}[t]
    \caption{Evaluation results of monocular 3D object detection on the KITTI-360 test set. $^{*}$Reproduced with the official code. $^{\dagger}$CAD models are used as extra data. $^{\ddagger}$$M(D)$ indicates that detection model $D$ is employed for model-agnostic method $M$.}
    \label{tab:experiments/evaluation_results/monocular_3d_object_detection/weakly_supervised_setting}
    \centering
    \begin{minipage}{\linewidth}
    \renewcommand{\footnoterule}{\empty}
    \renewcommand{\thefootnote}{\fnsymbol{footnote}}
    \renewcommand{\thempfootnote}{\fnsymbol{mpfootnote}}
    \scalebox{0.9}{
        \begin{tabular}{cccccccccccc}
            \toprule
            & \multicolumn{2}{c}{Weak Supervision} & Full Supervision & \multicolumn{2}{c}{AP$_{\text{BEV}}$/AP$_{\text{3D}}$@0.3} & \multicolumn{2}{c}{AP$_{\text{BEV}}$/AP$_{\text{3D}}$@0.5} \\
            \cmidrule(lr){2-3} \cmidrule(lr){4-4} \cmidrule(lr){5-6} \cmidrule(lr){7-8}
            Method                                                                                   & LiDAR  & Masks  & 3D Boxes & Easy        & Hard        & Easy        & Hard        \\
            \midrule
            WeakM3D\footnotemark[1] \cite{WeakM3D}                                                   & \cmark & \cmark &          & 29.89/21.25 & 24.01/15.34 & 8.10/2.96   & 2.96/2.01   \\
            Autolabels\footnotemark[1]\footnotemark[2]\footnotemark[3] \cite{Autolabels} (S-WeakM3D) & \cmark & \cmark &          & 48.16/12.92 & 37.34/9.94  & 20.18/4.69  & 14.33/2.79  \\
            VSRD\footnotemark[3] (S-WeakM3D)                                                         &        & \cmark &          & 51.09/42.94 & 41.28/33.78 & 19.51/11.91 & 14.39/8.46  \\
            VSRD\footnotemark[3] (MonoFlex)                                                          &        & \cmark &          & 54.40/48.16 & 45.67/40.04 & \textbf{29.18}/18.53 & 22.31/13.60 \\
            VSRD\footnotemark[3] (MonoDETR)                                                          &        & \cmark &          & \textbf{58.40}/\textbf{50.86} & \textbf{50.61}/\textbf{43.45} & 29.07/\textbf{21.77} & \textbf{22.83}/\textbf{16.46} \\
            \midrule
            MonoFlex\footnotemark[1] \cite{MonoFlex}                                                 &        &        & \cmark   & 69.70/67.07 & 59.86/57.26 & 50.82/43.11 & 41.78/34.43 \\
            MonoDETR\footnotemark[1] \cite{MonoDETR}                                                 &        &        & \cmark   & 63.07/60.49 & 54.04/50.03 & 47.21/41.01 & 36.05/30.38 \\
            \bottomrule
        \end{tabular}
    }
    \end{minipage}
    \vspace{-2mm}
\end{table*}

%% file: experiments/evaluation_results/monocular_3d_object_detection/semi_supervised_setting/table.tex
\begin{table}[t]
    \caption{Evaluation results of semi-supervised monocular 3D object detection on the KITTI validation set.}
    \label{tab:experiments/evaluation_results/monocular_3d_object_detection/semi_supervised_setting}
    \centering
    \setlength{\tabcolsep}{0.8mm}
    \scalebox{0.85}{
        \begin{tabular}{ccccc}
        \toprule
        & & \multicolumn{3}{c}{AP$_{\text{BEV}}$/AP$_{\text{3D}}$@0.7} \\
        \cmidrule(lr){3-5}        
        Method                           & Ratio & Easy                 & Moderate             & Hard                 \\
        \midrule
        MonoDETR \cite{MonoDETR}         & 1.00  & 37.99/29.36          & 26.76/20.64          & 23.02/17.30          \\ 
        \midrule
        \multirow{5}{*}{VSRD (MonoDETR)} & 0.00  & 0.002/0.001          & 0.004/0.001          & 0.005/0.002          \\
                                         & 0.25  & 31.72/21.76          & 22.32/15.43          & 18.86/12.55          \\
                                         & 0.50  & \textbf{43.44/31.05} & \textbf{31.54/21.48} & \textbf{27.17/17.93} \\
                                         & 0.75  & \textbf{42.58/32.95} & \textbf{31.08/24.68} & \textbf{27.19/21.38} \\
        \bottomrule
        \end{tabular}
    }
    \vspace{-4mm}
\end{table}

%% file: experiments/evaluation_results/multi_view_3d_auto_labeling/main.tex
\subsubsection{Multi-View 3D Auto-Labeling}
\label{sec:experiments/evaluation_results/multi_view_3d_auto_labeling}

Autolabels \cite{Autolabels} employs a similar two-stage framework consisting of auto-labeling and subsequent training of 3D object detectors using the pseudo labels. We compare our method with Autolabels to evaluate the quality of the pseudo labels. As our pseudo labels are unreliable for dynamic, occluded, or truncated objects, we group the dataset based on whether the average confidence score for each image is greater than a certain threshold. The evaluation results with the confidence thresholds of $0.0$ and $0.8$ are shown in \cref{tab:experiments/evaluation_results/multi_view_3d_auto_labeling}. Our method exhibits superior performance in terms of AP$_{\text{BEV}}@0.3$ and AP$_{\text{3D}}$. For AP${_\text{BEV}}@0.5$, our method exhibits slightly lower performance for the confidence threshold of $0.0$. However, by raising the confidence threshold to $0.8$, our method demonstrates superior performance, indicating that our method can generate more high-quality pseudo labels while abandoning low-quality pseudo labels. The optimized 3D bounding boxes and rendered instance masks are visualized in \cref{fig:experiments/evaluation_results/multi_view_3d_auto_labeling}.

%% file: experiments/evaluation_results/monocular_3d_object_detection/main.tex
\subsubsection{Monocular 3D Object Detection}
\label{sec:experiments/evaluation_results/monocular_3d_object_detection}

The pseudo labels generated by the proposed auto-labeling serve as 3D supervision. We further investigate their applicability using the existing monocular 3D object detectors.

\vspace{-3mm}
\input{experiments/evaluation_results/monocular_3d_object_detection/weakly_supervised_setting/main}
\vspace{-3mm}
\input{experiments/evaluation_results/monocular_3d_object_detection/semi_supervised_setting/main}

%% file: experiments/evaluation_results/monocular_3d_object_detection/weakly_supervised_setting/main.tex
\paragraph{Weakly Supervised Setting} 
\label{sec:experiments/evaluation_results/monocular_3d_object_detection/weakly_supervised_setting}

\cref{tab:experiments/evaluation_results/monocular_3d_object_detection/weakly_supervised_setting} shows the evaluation results of our method compared with the existing weakly supervised and fully supervised methods. Our method demonstrates a significant superiority over WeakM3D \cite{WeakM3D} across all the metrics while eliminating the need for LiDAR points for 3D supervision. Moreover, the detector trained on the pseudo labels generated by the proposed auto-labeling outperforms that trained on the pseudo labels generated by Autolabels \cite{Autolabels}. Furthermore, employing more sophisticated monocular 3D object detectors such as MonoFlex \cite{MonoFlex} and MonoDETR \cite{MonoDETR} further improves detection performance, demonstrating the broad versatility of our method, which is not limited to a specific detection model. It is noteworthy that the detectors trained on the pseudo labels generated by our method demonstrate competitive performance compared with those trained in a fully supervised manner.

%% file: experiments/evaluation_results/monocular_3d_object_detection/semi_supervised_setting/main.tex
\paragraph{Semi-Supervised Setting} 
\label{sec:experiments/evaluation_results/monocular_3d_object_detection/semi_supervised_setting}

The essential advantage of our method is that it avoids costly 3D annotations, making more data available for training. Therefore, we investigate a realistic scenario where a detector pre-trained on a large amount of unlabeled data from a source domain is fine-tuned on a small amount of labeled data from a target domain. We select the KITTI-360 and KITTI datasets as source and target domains, respectively. \cref{tab:experiments/evaluation_results/monocular_3d_object_detection/semi_supervised_setting} shows the performance of the detector pre-trained on the KITTI-360 dataset in a weakly supervised manner with the proposed auto-labeling and then fine-tuned on a subset of the KITTI dataset in a supervised manner. The zero-shot performance is quite low due to the characteristic that monocular depth estimation is greatly affected by the differences in camera parameters, but the performance of the detector fine-tuned on only 50\% of the labeled data significantly outperforms that trained on the whole data from scratch, highlighting the broad applicability of our method.

%% file: conclusion/main.tex
\section{Conclusion}
\label{sec:conclusion}

In this paper, we propose a novel weakly supervised 3D object detection framework named VSRD, which consists of multi-view 3D auto-labeling and subsequent training of monocular 3D object detectors using the pseudo labels generated in the auto-labeling stage. Our method demonstrates superior performance compared with the existing weakly supervised 3D object detection methods. Moreover, it exhibits remarkable scalability using partially labeled data for semi-supervised learning. Our proposed method allows leveraging abundant 2D annotations to enhance 3D object detection without explicit 3D supervision, providing a promising avenue for further advancements in the field.

%% file: additional_implementation_details/main.tex
\section{Additional Implementation Details}
\label{sec:additional_implementation_details}

\input{additional_implementation_details/multi_view_3d_auto_labeling/main}

\input{additional_implementation_details/monocular_3d_object_detection/main}

%% file: additional_implementation_details/multi_view_3d_auto_labeling/main.tex
\subsection{Multi-View 3D Auto-Labeling}
\label{sec:aditional_implementation_details/multi_view_3d_auto_labeling}

\input{additional_implementation_details/multi_view_3d_auto_labeling/cuboid_sdf/main}
\input{additional_implementation_details/multi_view_3d_auto_labeling/symmetric_shape_prior/main}
\input{additional_implementation_details/multi_view_3d_auto_labeling/frame_sampling/main}
\input{additional_implementation_details/multi_view_3d_auto_labeling/ray_sampling/main}

%% file: additional_implementation_details/multi_view_3d_auto_labeling/cuboid_sdf/main.tex
\subsubsection{Cuboid SDF}
\label{sec:additional_implementation_details/multi_view_3d_auto_labeling/cuboid_sdf}

SDFs for primitives such as spheres and cuboids can be derived theoretically. Here, we introduce the SDF for cuboids. First, we define the cuboid in a local coordinate system with a dimension $\bm{d} \in \mathbb{R}^{3}_{+}$ as the following set of vertices $\mathcal{C}_{\bm{d}}$:
\begin{align}
    \label{eq:additional_implementation_details/multi_view_3d_auto_labeling/cuboid_sdf/cuboid_vertices}
    \mathcal{C}_{\bm{d}} = \{-\frac{\bm{d}_{x}}{2}, \frac{\bm{d}_{x}}{2}\} \times \{-\frac{\bm{d}_{y}}{2}, \frac{\bm{d}_{y}}{2}\} \times \{-\frac{\bm{d}_{z}}{2}, \frac{\bm{d}_{z}}{2}\} \ .
\end{align}
Accordingly, the \textit{local} SDF $\bar{\mathcal{B}}(\cdot ; \bm{d})$ for the cuboid $\mathcal{C}_{\bm{d}}$ can be derived as follows:
\begin{align}
    \label{eq:additional_implementation_details/multi_view_3d_auto_labeling/cuboid_sdf/local_cuboid_sdf}
    \mathcal{\bar{B}}(\bm{p} ; \bm{d}) & = \| \max(\bm{q}, 0) \|_{2} + \min(m, 0) \ , \\
    \bm{q} & = |\bm{p}| - \bm{d} \ , \nonumber \\
    m & = \max(\bm{q}_{x}, \bm{q}_{y}, \bm{q}_{z}) \ . \nonumber
\end{align}
Next, we transform the cuboid SDF from the local coordinate system to the global one. In general, given a surface whose SDF is denoted by $\bar{\mathcal{F}}(\cdot)$, the SDF $\mathcal{F}(\cdot ; \bm{R}, \bm{t})$ for the surface transformed by a rigid transformation $(\bm{R}, \bm{t}) \in \textbf{SE}(3)$ is given by:
\begin{align}
    \label{eq:additional_implementation_details/multi_view_3d_auto_labeling/cuboid_sdf/rigid_transformation}
    \mathcal{F}(\bm{p} ; \bm{R}, \bm{t}) = \bar{\mathcal{F}}(\bm{R}^T (\bm{p} - \bm{t})) \ .
\end{align}
Therefore, the \textit{global} cuboid SDF $\mathcal{B}(\cdot ; \bm{d}, \bm{\ell}, \bm{R})$ parameterized by a dimension $\bm{d} \in \mathbb{R}^{3}_{+}$, location $\bm{\ell} \in \mathbb{R}^{3}$, and orientation $\bm{R} \in \text{SO}(3)$ can be derived by transforming the \textit{local} cuboid SDF $\mathcal{\bar{B}}(\cdot ; \bm{d})$ from the local coordinate system to the global one with the rigid transformation $(\bm{R}, \bm{\ell})$, as follows:
\begin{align}
	\label{eq:additional_implementation_details/multi_view_3d_auto_labeling/cuboid_sdf/global_cuboid_sdf}
	\mathcal{B}(\bm{p} ; \bm{d}, \bm{\ell}, \bm{R}) = \mathcal{\bar{B}}(\bm{R}^{T} (\bm{p} - \bm{\ell}) ; \bm{d}) \ .
\end{align} 

%% file: additional_implementation_details/multi_view_3d_auto_labeling/symmetric_shape_prior/main.tex
\subsubsection{Symmetric Shape Prior}
\label{sec:aditional_implementation_details/multi_view_3d_auto_labeling/symmetric_shape_prior}

Since the shapes of vehicles are often horizontally symmetrical in the local coordinate system, we incorporate this shape prior to each instance SDF in our proposed multi-view 3D auto-labeling. Given a \textit{local} SDF $\mathcal{F}(\cdot)$, its horizontally symmetrical version $\overleftrightarrow{\mathcal{F}}(\cdot)$ is given by:
\begin{align}
    \label{eq:aditional_implementation_details/multi_view_3d_auto_labeling/symmetric_shape_prior/symmetric_local_sdf}
    \overleftrightarrow{\mathcal{F}}(\bm{p}) & = \mathcal{F}([|\bm{p}_{x}|, \bm{p}_{y}, \bm{p}_{z}]) \ . 
\end{align}
This symmetric shape prior has the advantage that even if only one of the left or right sides of an instance is visible, the shape of the invisible part can be shared with the other instances via the hypernetwork.

%% file: additional_implementation_details/multi_view_3d_auto_labeling/frame_sampling/main.tex
\subsubsection{Frame Sampling}
\label{sec:additional_implementation_details/multi_view_3d_auto_labeling/frame_sampling}

Since our loss functions are based on multi-view 2D supervision, how to sample source frames $\mathcal{S}$ for each target frame $t$ is important. To sample source frames, each of which includes as many instances in the target frame as possible, we sample source frames based on the perspective of \textit{what percentage of instances in the target frame are included in each source frame}. Therefore, we first define a set of candidate source frames $\tilde{\mathcal{S}}(\eta)$ as the set of frames with the maximum number of elements, where the percentage of \textit{target} instances in every source frame is greater than or equal to a certain threshold $\eta$ as follows:
\begin{align}
    \label{eq:additional_implementation_details/multi_view_3d_auto_labeling/frame_sampling/source_frames}
    \tilde{\mathcal{S}}(\eta) = \max_{|\cdot|} (\{\mathcal{N} \ni t \mid \forall s \in \mathcal{N} : \frac{|\mathcal{I}_{t} \cap \mathcal{I}_{s}|}{|\mathcal{I}_{t}|} \ge \eta\}) \ , 
\end{align}
where $\mathcal{I}_{t}$ and $\mathcal{I}_{s}$ denote the sets of instance IDs of the target and source frames, respectively. $\max_{|\cdot|}$ denotes the \textit{max} operation that selects the set with the maximum number of elements. In our experiments, we set empirically $\eta = 0.5$, balancing the number of viewpoints and convergence time. In practice, we further sample a fixed number of frames from $\tilde{\mathcal{S}}(\eta)$ as evenly as possible due to implementation considerations to avoid large differences in the number of source frames between scenes. We sample $16$ frames from $\tilde{\mathcal{S}}(0.5)$ and use it as the final set of source frames $\mathcal{S}$.

%% file: additional_implementation_details/multi_view_3d_auto_labeling/ray_sampling/main.tex
\subsubsection{Ray Sampling}
\label{sec:additional_impementation_details/multi_view_3d_auto_labeling/ray_sampling}

For each iteration in stochastic gradient descent, as in NeRF \cite{NeRF}, we randomly sample a batch of rays used for volumetric rendering. However, in the case where the instance masks are given, it is inefficient to sample rays far away from any instance in the scene. Therefore, we propose an efficient ray sampling algorithm based on the instance masks. First, for each source frame $i \in \mathcal{S}$, we extract the polygon for each instance by a contour finder. Then, we theoretically derive the 2D SDF $\mathcal{P}_{in}(\cdot)$ for the $n$-th polygon in frame $i$. Then, we derive the 2D SDF $\mathcal{P}_{i}(\cdot)$ for the union of all the polygons in frame $i$ as $\mathcal{P}_{i}(\bm{p}) = \min(\mathcal{P}_{i1}(\bm{p}), \ldots, \mathcal{P}_{iN_{i}}(\bm{p}))$, where $N_{i}$ denotes the number of \textit{target} instances in frame $i$. Then, we generate a \textit{soft} instance mask $M_{i}$ via \textit{soft rasterization} \cite{SoftRasterizer} as follows:
\begin{align}
    \label{eq:additional_impementation_details/multi_view_3d_auto_labeling/ray_sampling/soft_instance_mask}
    M_{i}(\bm{p}) = \Phi(-\mathcal{P}_{i}(\bm{p}) / \tau) \ , \
\end{align}
where $\bm{p} \in \mathbb{R}^{2}$ denotes a pixel coordinate, $\Phi(
\cdot)$ denotes the Sigmoid function, and $\tau$ denotes the temperature parameter that controls the degree of relaxation, indicating that as it becomes higher, rays farther away from each instance are sampled. Then, we normalize the soft instance mask $M_{i}$ across all the source frames $\mathcal{S}$ as follows:
\begin{align}
    \label{eq:additional_impementation_details/multi_view_3d_auto_labeling/ray_sampling/normalized_soft_instance_mask}
    \tilde{M}_{i}(\bm{p}) = \frac{M_{i}(\bm{p})}{\sum_{j \in \mathcal{S}} \sum_{\bm{q}} M_{j}(\bm{q})} \ . 
\end{align}
Finally, we sample a batch of rays from the multinomial distribution based on the normalized soft instance mask $\tilde{M}_{i}(\cdot)$. This mechanism enables us to intensively sample rays that are likely to hit the surface of each instance.

%% file: additional_implementation_details/monocular_3d_object_detection/main.tex
\subsection{Monocular 3D Object Detection}
\label{sec:additional_implementation_details/monocular_3d_object_detection}

To compare our method and Autolabels \cite{Autolabels} with WeakM3D \cite{WeakM3D}, we modify the architecture of WeakM3D so that it can be trained in a supervised manner using pseudo labels. More specifically, we train dimension and confidence heads in addition to the existing location and orientation heads using the same supervised loss as MonoDIS \cite{MonoDIS}. We call this model S-WeakM3D. Note that for the comparison with methods other than WeakM3D, we do not modify the architectures and loss functions.

\input{additional_implementation_details/monocular_3d_object_detection/architecture/main}

\input{additional_implementation_details/monocular_3d_object_detection/loss_functions/main}

\input{additional_evaluation_results/monocular_3d_object_detection/weakly_supervised_setting/table}
\input{additional_evaluation_results/monocular_3d_object_detection/semi_supervised_setting/table}

%% file: additional_implementation_details/monocular_3d_object_detection/architecture/main.tex
\subsubsection{Architecture}
\label{sec:additional_implementation_details/monocular_3d_object_detection/architecture}

\input{additional_implementation_details/monocular_3d_object_detection/architecture/dimension_head/main}
\input{additional_implementation_details/monocular_3d_object_detection/architecture/confidence_head/main}

%% file: additional_implementation_details/monocular_3d_object_detection/architecture/dimension_head/main.tex
\paragraph{Dimension Head}
\label{sec:additional_implementation_details/monocular_3d_object_detection/architecture/dimension_head}

WeakM3D utilizes prior knowledge about the typical dimension for category \textit{Car} and freezes it by setting the width, height, and length to $1.8$, $1.6$, and $4.0$, respectively. This is because relying only on the 3D alignment loss utilizing LiDAR point clouds struggles to optimize the location and dimension parameters jointly. In order to make full use of the pseudo labels generated by the proposed auto-labeling, we modify the original network of WeakM3D by adding a simple dimension head, which has the same architecture as the original location head, to estimate the dimension of each instance as follows:
\begin{align}
    \label{sec:additional_implementation_details/monocular_3d_object_detection/architecture/dimension_head/dimension_head}
    \hat{\bm{d}} = \bm{d}_{\text{min}} + (\bm{d}_{\text{max}} - \bm{d}_{\text{min}}) \odot \Phi(\mathcal{H}_{\text{dim}}(F ; \bm{\psi}_{\text{dim}})) \ ,
\end{align}
where $\bm{d}_{\text{min}} = [1.5, 1.5, 3.0]$ and $\bm{d}_{\text{max}} = [2.0, 2.0, 5.0]$ denote the pre-defined minimum and maximum dimensions, respectively. As we assume we cannot access any 3D bounding boxes in the dataset, they are determined based on not the statistics of the dataset but the dimensions of typical production cars. $F$ denotes the RoI-aligned feature maps for each instance, $\mathcal{H}_{\text{dim}}(\cdot ; \bm{\psi}_{\text{dim}})$ denotes the dimension head parameterized by $\bm{\psi}_{\text{dim}}$, and $\Phi(\cdot)$ denotes the Sigmoid function. We implement the dimension head as an MLP with two hidden layers, each of which has $256$ channels. 

%% file: additional_implementation_details/monocular_3d_object_detection/architecture/confidence_head/main.tex
\paragraph{Confidence Head}
\label{sec:additional_implementation_details/monocular_3d_object_detection/architecture/confidence_head}

Following MonoDIS \cite{MonoDIS}, we train the network to estimate not only the 3D bounding box but also a confidence score that represents the quality of the predicted 3D bounding box in a self-supervised manner. We add a simple confidence head, which has the same architecture as the original location head, to estimate the confidence of the predicted 3D bounding box as follows:
\begin{align}
     \label{eq:additional_implementation_details/monocular_3d_object_detection/architecture/confidence_head/confidence_head}
     \ \hat{c} = \Phi(\mathcal{H}_{\text{conf}}(F ; \bm{\psi}_{\text{conf}})) \ ,
\end{align}
where $\mathcal{H}_{\text{conf}}(\cdot ; \bm{\psi}_{\text{conf}})$ denotes the confidence head parameterized by $\bm{\psi}_{\text{conf}}$. We implement the confidence head as an MLP with two hidden layers, each of which has $256$ channels. We train the confidence head with the self-supervised loss explained in \cref{sec:additional_implementation_details/monocular_3d_object_detection/loss_functions}. The output confidence score is further multiplied by the classification score output by the off-the-shelf 2D detector and used as the final score to filter low-quality predictions during inference.

%% file: additional_implementation_details/monocular_3d_object_detection/loss_functions/main.tex
\subsubsection{Loss Functions}
\label{sec:additional_implementation_details/monocular_3d_object_detection/loss_functions}

\input{additional_implementation_details/monocular_3d_object_detection/loss_functions/disentangled_loss/main}
\input{additional_implementation_details/monocular_3d_object_detection/loss_functions/confidence_loss/main}

%% file: additional_implementation_details/monocular_3d_object_detection/loss_functions/disentangled_loss/main.tex
\paragraph{Distentangled loss} 
\label{sec:additional_implementation_details/monocular_3d_object_detection/loss_functions/disentangled_loss}

For bounding box regression, we employ the same disentangled loss as MonoDIS \cite{MonoDIS} as follows: 
\begin{align}
	\label{eq:additional_implementation_details/monocular_3d_object_detection/loss_functions/disentangled_loss/disentangled_loss}
	\mathcal{L}_{\text{box}}(\hat{\bm{d}}, \hat{\bm{l}}, \hat{\theta}, \bm{d}, \bm{l}, \theta) = \ 
	& \| B(\hat{\bm{d}}, \bm{l}, \theta) - B(\bm{d}, \bm{l}, \theta) \|_{\rm{H}} + \nonumber \\ 
	& \| B(\bm{d}, \hat{\bm{l}}, \theta) - B(\bm{d}, \bm{l}, \theta) \|_{\rm{H}} + \nonumber \\
	& \| B(\bm{d}, \bm{l}, \hat{\theta}) - B(\bm{d}, \bm{l}, \theta) \|_{\rm{H}} \ ,
\end{align}
where $\hat{\bm{d}}$, $\hat{\bm{l}}$, and $\hat{\theta}$ denote the predicted dimension, location, and orientation, respectively, and $\bm{d}$, $\bm{l}$, and $\theta$ denote the ground truth dimension, location, and orientation, respectively. $B(\bm{d}, \bm{l}, \theta)$ denotes the 3D bounding box decoded from dimension $\bm{d}$, location $\bm{l}$, and orientation $\theta$. Actually, we minimize the confidence-based weighted regression loss $\tilde{\mathcal{L}}_{\text{box}}$ instead of the original regression loss $\mathcal{L}_{\text{box}}$, as explained in Sec. \textcolor{red}{3.2.2} in the main paper.

%% file: additional_implementation_details/monocular_3d_object_detection/loss_functions/confidence_loss/main.tex
\paragraph{Confidence Loss}
\label{sec:additional_implementation_details/monocular_3d_object_detection/loss_functions/confidence_loss}

For confidence learning, we employ the same self-supervised loss as MonoDIS \cite{MonoDIS} as follows: 
\begin{align}
	\label{eq:additional_implementation_details/monocular_3d_object_detection/loss_functions/confidence_loss/confidence_loss}
	\mathcal{L}_{\text{conf}}(\hat{c}, \hat{\bm{d}}, \hat{\bm{l}}, \hat{\theta}, \bm{d}, \bm{l}, \theta) = \text{BCE}(\hat{c}, c) \ , \\
	c = \exp(- \lfloor \mathcal{L}_{\text{box}}(\hat{\bm{d}}, \hat{\bm{l}}, \hat{\theta}, \bm{d}, \bm{l}, \theta) \rfloor) \ , 
\end{align}
where $\hat{c}$ denotes the predicted confidence, $\text{BCE}(\cdot, \cdot)$ denotes the binary cross entropy, and $\lfloor \cdot \rfloor$ denotes the \textit{stop gradient} operation whereby the gradients are not propagated through the box regression loss $\mathcal{L}_{\text{box}}$.

%% file: additional_evaluation_results/monocular_3d_object_detection/weakly_supervised_setting/table.tex
\begin{table*}[t]
    \caption{Evaluation results of monocular 3D object detection on the KITTI-360 validation set. $^{*}$Reproduced with the official code. $^{\dagger}$CAD models are used as extra data. $^{\ddagger}$$M(D)$ indicates that detection model $D$ is employed for model-agnostic method $M$.}
    \label{tab:additional_evaluation_results/monocular_3d_object_detection/weakly_supervised_setting}
    \centering
    \begin{minipage}{\linewidth}
    \renewcommand{\footnoterule}{\empty}
    \renewcommand{\thefootnote}{\fnsymbol{footnote}}
    \renewcommand{\thempfootnote}{\fnsymbol{mpfootnote}}
    \scalebox{0.9}{
        \begin{tabular}{cccccccccccc}
            \toprule
            & \multicolumn{2}{c}{Weak Supervision} & Full Supervision & \multicolumn{2}{c}{AP$_{\text{BEV}}$/AP$_{\text{3D}}$@0.3} & \multicolumn{2}{c}{AP$_{\text{BEV}}$/AP$_{\text{3D}}$@0.5} \\
            \cmidrule(lr){2-3} \cmidrule(lr){4-4} \cmidrule(lr){5-6} \cmidrule(lr){7-8}
            Method                                                                                   & LiDAR  & Masks  & 3D Boxes & Easy        & Hard        & Easy        & Hard        \\
            \midrule
            WeakM3D\footnotemark[1] \cite{WeakM3D}                                                   & \cmark & \cmark &          & 49.38/44.26 & 41.53/34.91 & 17.25/4.64  & 13.87/3.45  \\
            Autolabels\footnotemark[1]\footnotemark[2]\footnotemark[3] \cite{Autolabels} (S-WeakM3D) & \cmark & \cmark &          & 55.55/10.04 & 51.59/8.50  & 36.06/1.56  & 28.12/1.13  \\
            VSRD\footnotemark[3] (S-WeakM3D)                                                         &        & \cmark &          & 62.77/57.28 & 57.35/51.79 & 31.84/29.50 & 30.04/24.93 \\
            VSRD\footnotemark[3] (MonoFlex)                                                          &        & \cmark &          & \textbf{70.04/65.09} & \textbf{60.53/55.75} & \textbf{50.59/32.52} & \textbf{48.83/25.70} \\
            VSRD\footnotemark[3] (MonoDETR)                                                          &        & \cmark &          & 54.97/50.13 & 49.81/46.13 & 38.09/29.52 & 31.68/24.25 \\
            \midrule
            MonoFlex\footnotemark[1] \cite{MonoFlex}                                                 &        &        & \cmark   & 80.47/78.15 & 72.97/68.74 & 66.81/60.46 & 57.37/49.38 \\
            MonoDETR\footnotemark[1] \cite{MonoDETR}                                                 &        &        & \cmark   & 73.21/72.67 & 68.58/66.27 & 61.47/58.35 & 54.53/49.91 \\
            \bottomrule
        \end{tabular}
    }
    \end{minipage}
    \vspace{-2mm}
\end{table*}

%% file: additional_evaluation_results/monocular_3d_object_detection/semi_supervised_setting/table.tex
\begin{table}[t]
    \caption{Evaluation results of semi-supervised monocular 3D object detection on the KITTI validation set.}
    \label{tab:additional_evaluation_results/monocular_3d_object_detection/semi_supervised_setting}
    \centering
    \setlength{\tabcolsep}{0.8mm}
    \scalebox{0.85}{
        \begin{tabular}{ccccc}
        \toprule
        & & \multicolumn{3}{c}{AP$_{\text{BEV}}$/AP$_{\text{3D}}$@0.7} \\
        \cmidrule(lr){3-5}        
        Method                           & Ratio & Easy                 & Moderate             & Hard                 \\
        \midrule
        MonoFlex \cite{MonoFlex}         & 1.00  & 28.17/23.64          & 21.92/17.51          & 19.07/14.83          \\ 
        \midrule
        \multirow{5}{*}{VSRD (MonoFlex)} & 0.00  & 3.65/0.34            & 2.51/0.23            & 1.98/0.21            \\
                                         & 0.25  & 24.55/14.62          & 17.74/10.69          & 15.67/8.95           \\
                                         & 0.50  & \textbf{29.38}/17.44 & 21.72/12.40          & 18.69/10.65          \\
                                         & 0.75  & \textbf{34.32/23.79} & \textbf{24.87/17.60} & \textbf{21.45/14.97} \\
        \bottomrule
        \end{tabular}
    }
    \vspace{-4mm}
\end{table}

%% file: additional_evaluation_results/main.tex
\section{Additional Evaluation Results}
\label{sec:additional_evaluation_results}

\input{additional_evaluation_results/monocular_3d_object_detection/main}

%% file: additional_evaluation_results/monocular_3d_object_detection/main.tex
\subsection{Monocular 3D Object Detection}
\label{sec:additional_evaluation_results/monocular_3d_object_detection}

\input{additional_evaluation_results/monocular_3d_object_detection/weakly_supervised_setting/main}
\input{additional_evaluation_results/monocular_3d_object_detection/semi_supervised_setting/main}

%% file: additional_evaluation_results/monocular_3d_object_detection/weakly_supervised_setting/main.tex
\subsubsection{Weakly Supervised Setting}
\label{sec:additional_evaluation_results/monocular_3d_object_detection/weakly_supervised_setting}

\cref{tab:additional_evaluation_results/monocular_3d_object_detection/weakly_supervised_setting} shows the additional evaluation results of our method compared with the existing weakly supervised and fully supervised methods. As with Tab. \textcolor{red}{4} in the main paper, our method demonstrates a significant superiority over WeakM3D \cite{WeakM3D} across all the metrics while eliminating the need for LiDAR points for 3D supervision. Moreover, the detector trained on the pseudo labels generated by the proposed auto-labeling outperforms that trained on the pseudo labels generated by Autolabels \cite{Autolabels}. 

%% file: additional_evaluation_results/monocular_3d_object_detection/semi_supervised_setting/main.tex
\subsubsection{Semi-Supervised Setting}
\label{sec:additional_evaluation_results/monocular_3d_object_detection/semi_supervised_setting}

In addition to MonoDETR \cite{MonoDETR}, we also conduct the same experiments as Sec. \textcolor{red}{4.4.2} in the main paper employing MonoFlex \cite{MonoFlex}. \cref{tab:additional_evaluation_results/monocular_3d_object_detection/semi_supervised_setting} shows the performance of the detector pre-trained on the KITTI-360 dataset in a weakly supervised manner with the proposed auto-labeling and then fine-tuned on a subset of the KITTI dataset in a supervised manner. As with Tab. \textcolor{red}{5} in the main paper, the zero-shot performance is quite low due to the characteristic that monocular depth estimation is greatly affected by the differences in camera parameters, but the performance of the detector fine-tuned on only 75\% of the labeled data significantly outperforms that trained on the whole data from scratch, highlighting the broad applicability of our method.

%% file: additional_visualization_results/main.tex
\section{Additional Visualization Results}
\label{sec:additional_visualization_results}

\cref{fig:additional_visualization_results/multi_view_3d_auto_labeling,fig:additional_visualization_results/monocular_3d_object_detection/weakly_supervised_setting,fig:additional_visualization_results/monocular_3d_object_detection/semi_supervised_setting/monodetr,fig:additional_visualization_results/monocular_3d_object_detection/semi_supervised_setting/monoflex} show the additional visualization results of the proposed multi-view 3D auto-labeling, weakly supervised monocular 3D object detection, and semi-supervised monocular 3D object detection employing MonoDETR \cite{MonoDETR} and MonoFlex \cite{MonoFlex}, respectively. In particular, as can be seen from \cref{fig:additional_visualization_results/monocular_3d_object_detection/weakly_supervised_setting}, it is worth noting that WeakM3D \cite{WeakM3D} struggles to estimate the orientations of laterally moving objects accurately as it assumes that most objects are facing forward, whereas our method leverages the pseudo labels generated by the proposed auto-labeling as 3D supervision without any priors, leading to more accurate orientation estimation.



\input{additional_visualization_results/multi_view_3d_auto_labeling/figure}
\input{additional_visualization_results/monocular_3d_object_detection/weakly_supervised_setting/figure}
\clearpage
\input{additional_visualization_results/monocular_3d_object_detection/semi_supervised_setting/figure}

%% file: additional_visualization_results/multi_view_3d_auto_labeling/figure.tex
\begin{figure*}[!t]
    \begin{minipage}[b]{0.5\linewidth}
        \centering
        \includegraphics[width=1.0\linewidth]{figures/silhouette/0000003451.jpg}
    \end{minipage}
    \begin{minipage}[b]{0.5\linewidth}
        \centering
        \includegraphics[width=1.0\linewidth]{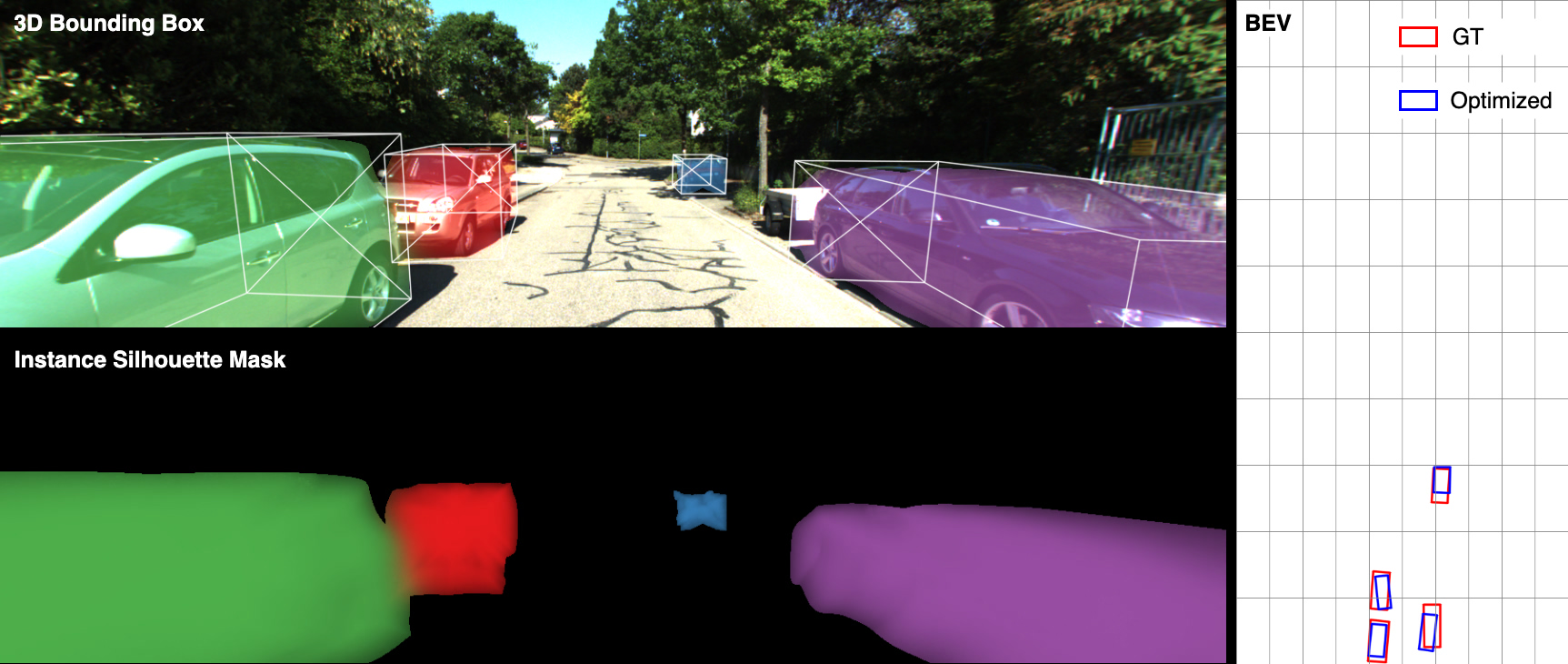}
    \end{minipage}
    \begin{minipage}[b]{0.5\linewidth}
        \centering
        \includegraphics[width=1.0\linewidth]{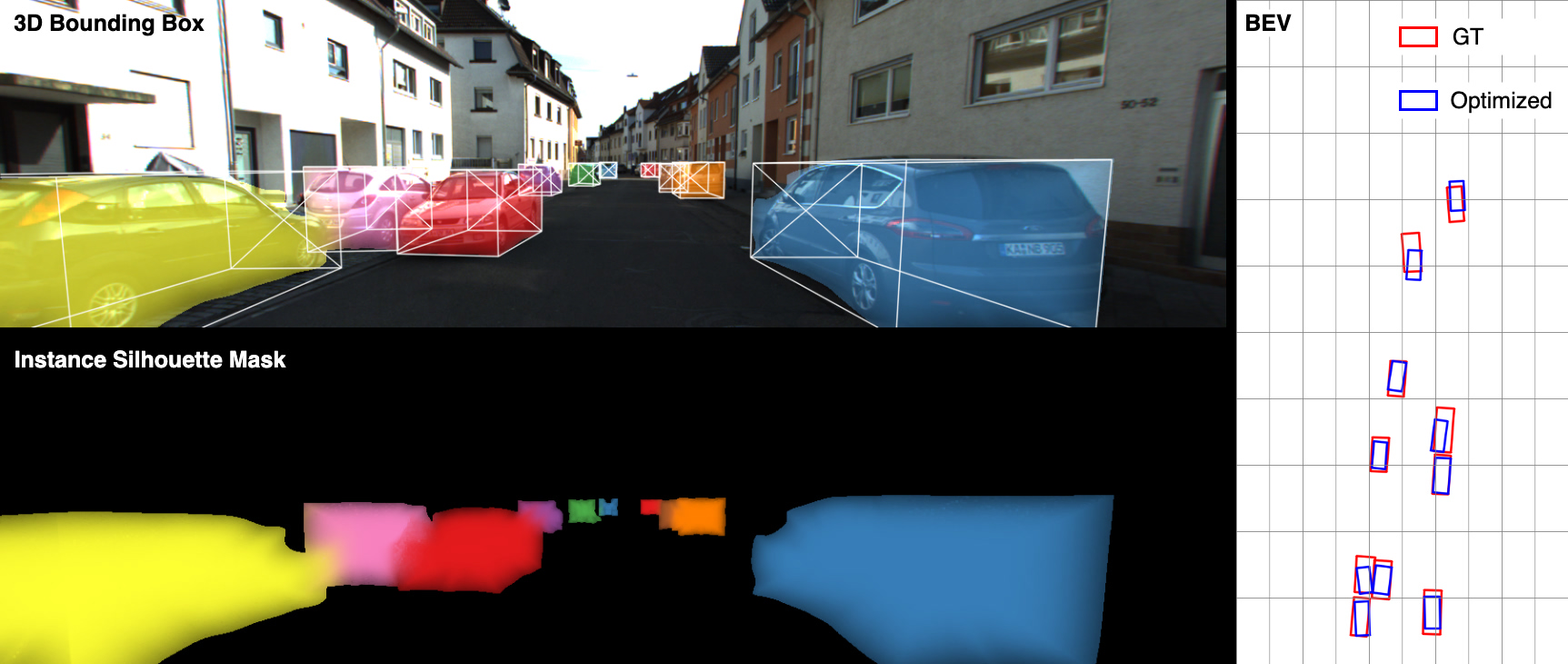}
    \end{minipage}
    \begin{minipage}[b]{0.5\linewidth}
        \centering
        \includegraphics[width=1.0\linewidth]{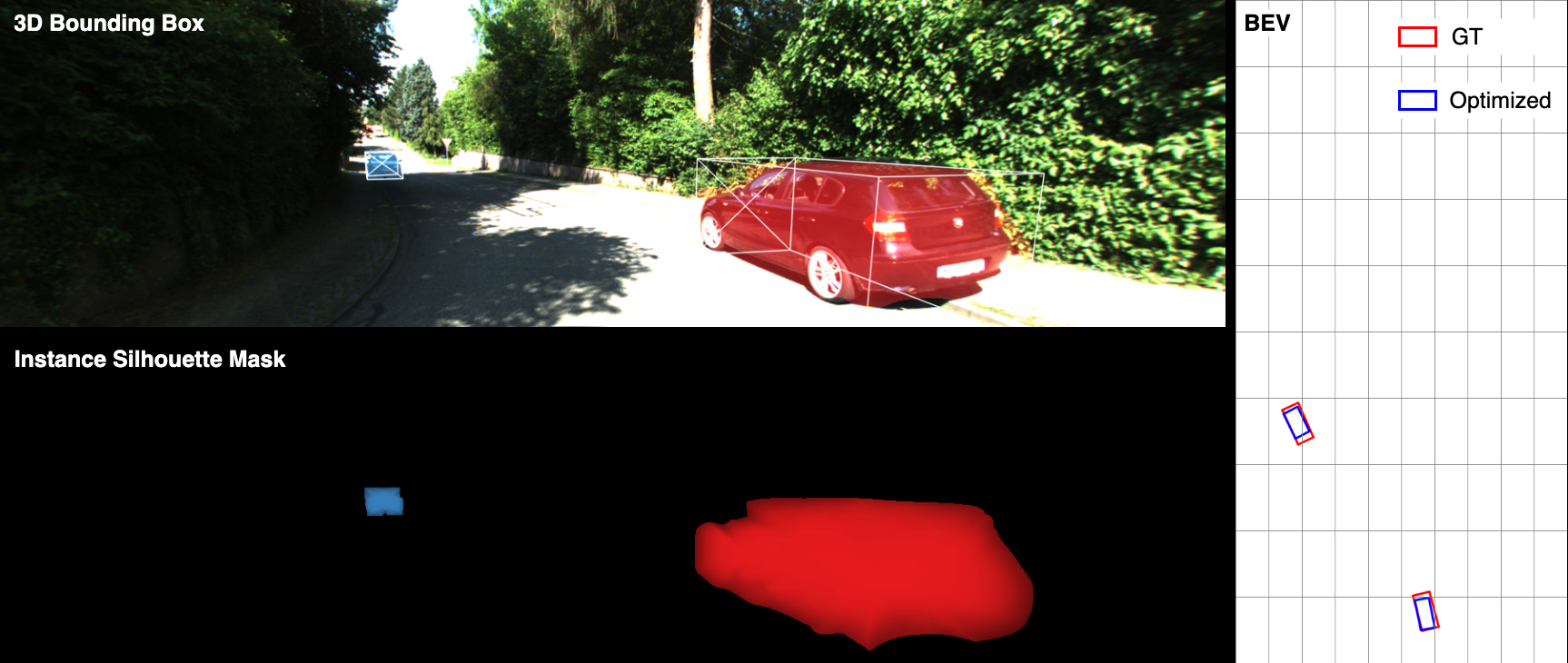}
    \end{minipage}
    \caption{Visualization results of the optimized 3D bounding boxes (1st row) and rendered instance masks (2nd row). We assign a unique color to each instance, and each pixel is colored as the weighted summation based on the rendered soft instance label.}
    \label{fig:additional_visualization_results/multi_view_3d_auto_labeling}
\end{figure*}

%% file: additional_visualization_results/monocular_3d_object_detection/weakly_supervised_setting/figure.tex
\begin{figure*}[t]
  \begin{minipage}[b]{0.12\linewidth}
    \centering
    \subcaption{WeakM3D}
    \vspace{5mm}
  \end{minipage}
  \begin{minipage}[b]{0.44\linewidth}
    \centering
    \includegraphics[width=1.0\linewidth]{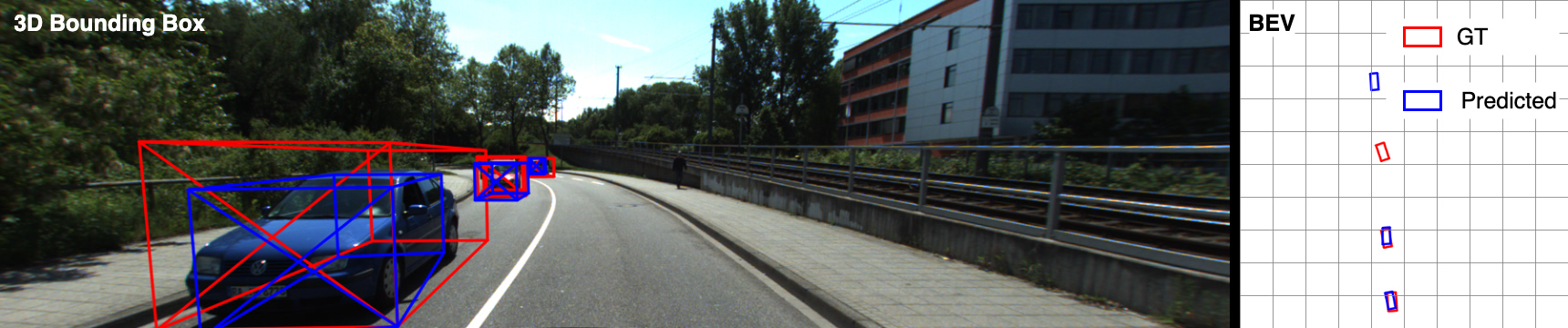}
  \end{minipage}
  \begin{minipage}[b]{0.44\linewidth}
    \centering
    \includegraphics[width=1.0\linewidth]{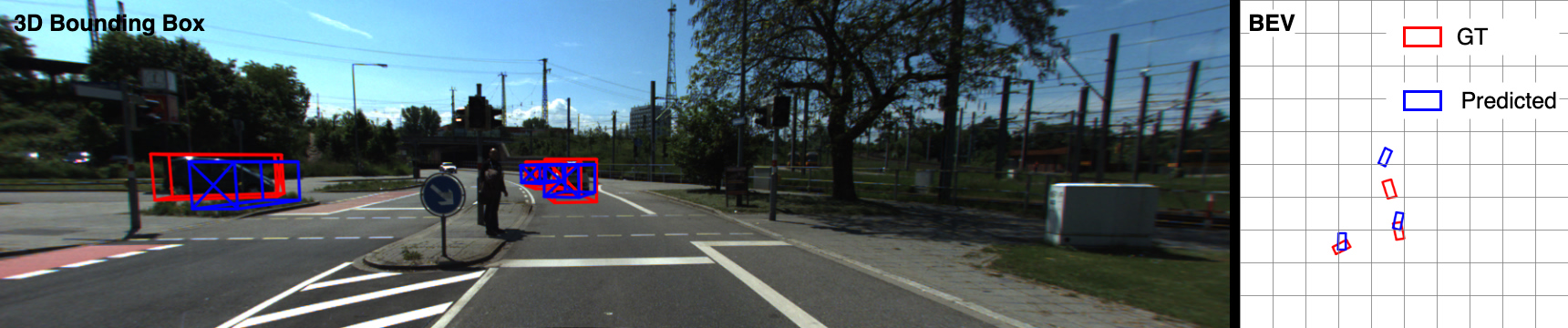}
  \end{minipage}
  \begin{minipage}[b]{0.12\linewidth}
    \centering
    \subcaption{Autolabels}
    \vspace{5mm}
  \end{minipage}
  \begin{minipage}[b]{0.44\linewidth}
    \centering
    \includegraphics[width=1.0\linewidth]{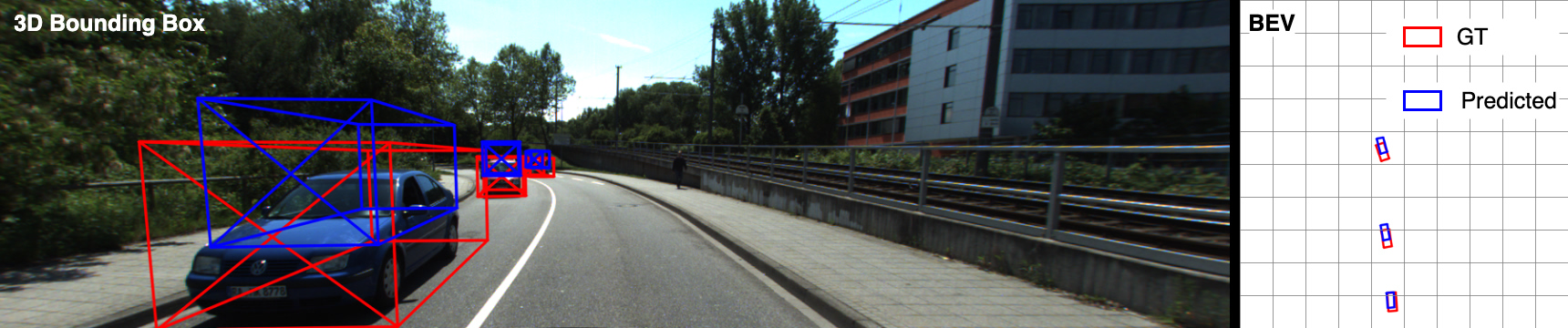}
  \end{minipage}
  \begin{minipage}[b]{0.44\linewidth}
    \centering
    \includegraphics[width=1.0\linewidth]{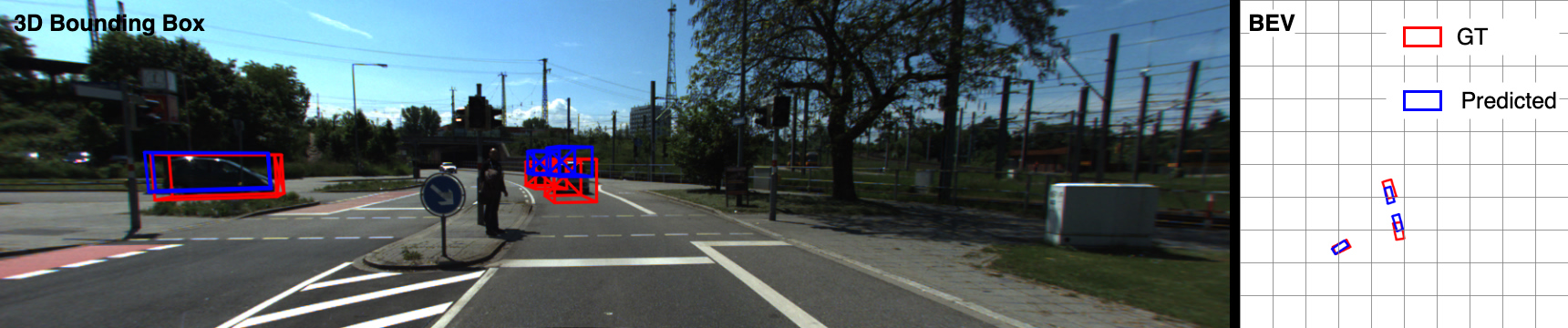}
  \end{minipage}
  \begin{minipage}[b]{0.12\linewidth}
    \centering
    \subcaption{VSRD}
    \vspace{5mm}
  \end{minipage}
  \begin{minipage}[b]{0.44\linewidth}
    \centering
    \includegraphics[width=1.0\linewidth]{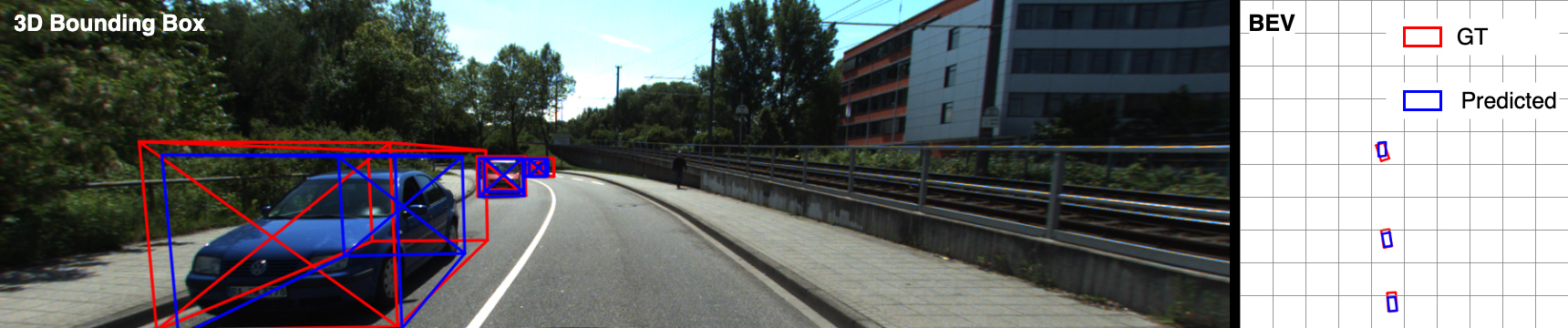}
  \end{minipage}
  \begin{minipage}[b]{0.44\linewidth}
    \centering
    \includegraphics[width=1.0\linewidth]{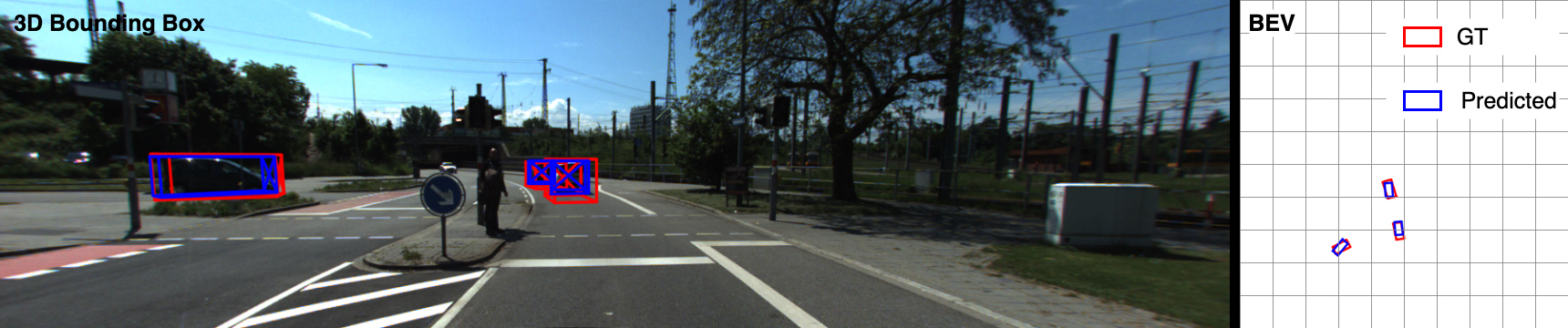}
  \end{minipage}
  \caption{Visualization results of weakly supervised monocular 3D object detection compared with Autolabels \cite{Autolabels} and WeakM3D \cite{WeakM3D}. The ground truth and predicted bounding boxes are drawn in \textcolor{red}{red} and \textcolor{blue}{blue}, respectively.}
  \label{fig:additional_visualization_results/monocular_3d_object_detection/weakly_supervised_setting}
\end{figure*}

%% file: additional_visualization_results/monocular_3d_object_detection/semi_supervised_setting/figure.tex
\begin{figure*}[t]
  \begin{minipage}[b]{0.12\linewidth}
    \centering
    \subcaption{MonoDETR}
    \vspace{5mm}
  \end{minipage}
  \begin{minipage}[b]{0.44\linewidth}
    \centering
    \includegraphics[width=1.0\linewidth]{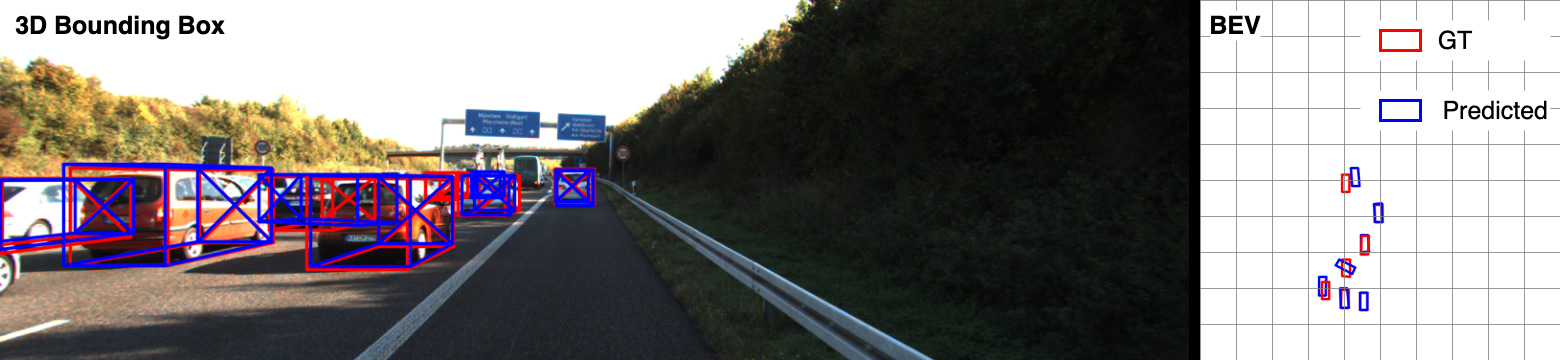}
  \end{minipage}
  \begin{minipage}[b]{0.44\linewidth}
    \centering
    \includegraphics[width=1.0\linewidth]{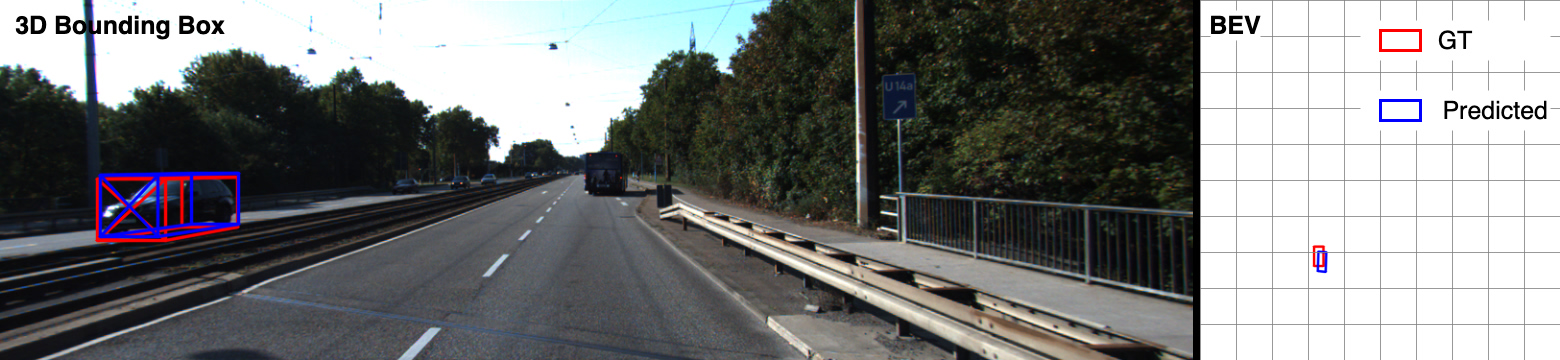}
  \end{minipage}
  \begin{minipage}[b]{0.12\linewidth}
    \centering
    \subcaption{VSRD (50\%)}
    \vspace{5mm}
  \end{minipage}
  \begin{minipage}[b]{0.44\linewidth}
    \centering
    \includegraphics[width=1.0\linewidth]{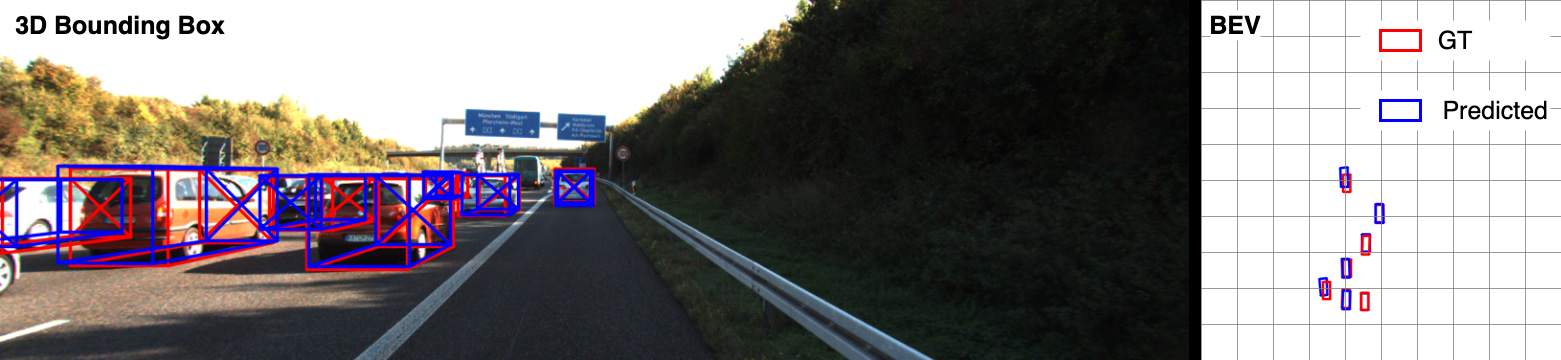}
  \end{minipage}
  \begin{minipage}[b]{0.44\linewidth}
    \centering
    \includegraphics[width=1.0\linewidth]{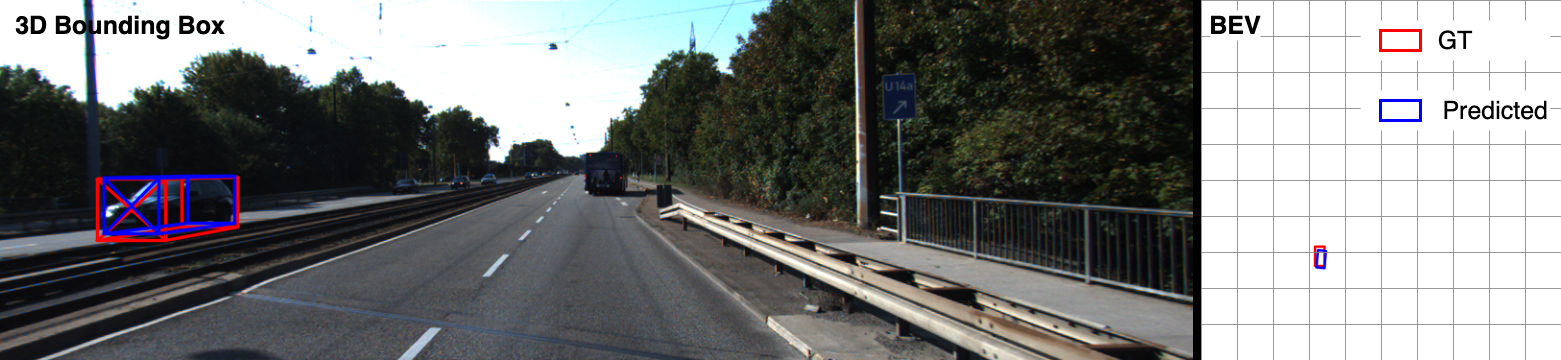}
  \end{minipage}
  \begin{minipage}[b]{0.12\linewidth}
    \centering
    \subcaption{VSRD (75\%)}
    \vspace{5mm}
  \end{minipage}
  \begin{minipage}[b]{0.44\linewidth}
    \centering
    \includegraphics[width=1.0\linewidth]{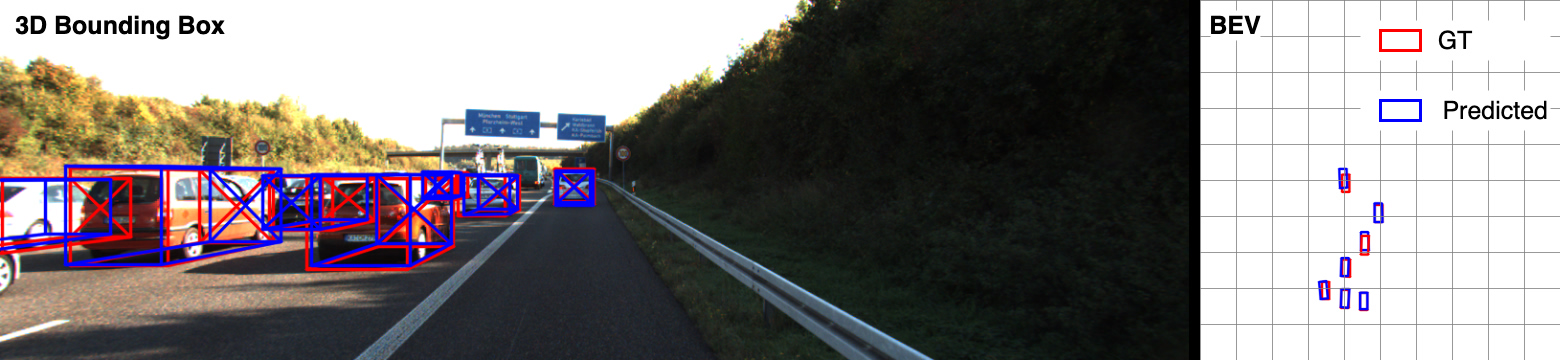}
  \end{minipage}
  \begin{minipage}[b]{0.44\linewidth}
    \centering
    \includegraphics[width=1.0\linewidth]{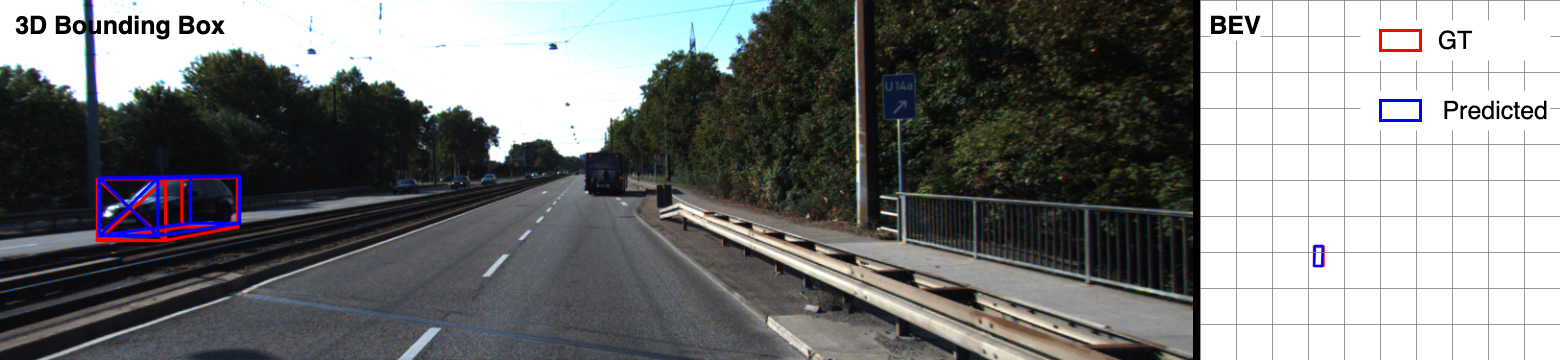}
  \end{minipage}
  \caption{Visualization results of semi-supervised monocular 3D object detection compared with MonoDETR \cite{MonoDETR}. The ground truth and predicted bounding boxes are drawn in \textcolor{red}{red} and \textcolor{blue}{blue}, respectively.}
  \label{fig:additional_visualization_results/monocular_3d_object_detection/semi_supervised_setting/monodetr}
\end{figure*}

\begin{figure*}[!t]
  \begin{minipage}[b]{0.12\linewidth}
    \centering
    \subcaption{MonoFlex}
    \vspace{5mm}
  \end{minipage}
  \begin{minipage}[b]{0.44\linewidth}
    \centering
    \includegraphics[width=1.0\linewidth]{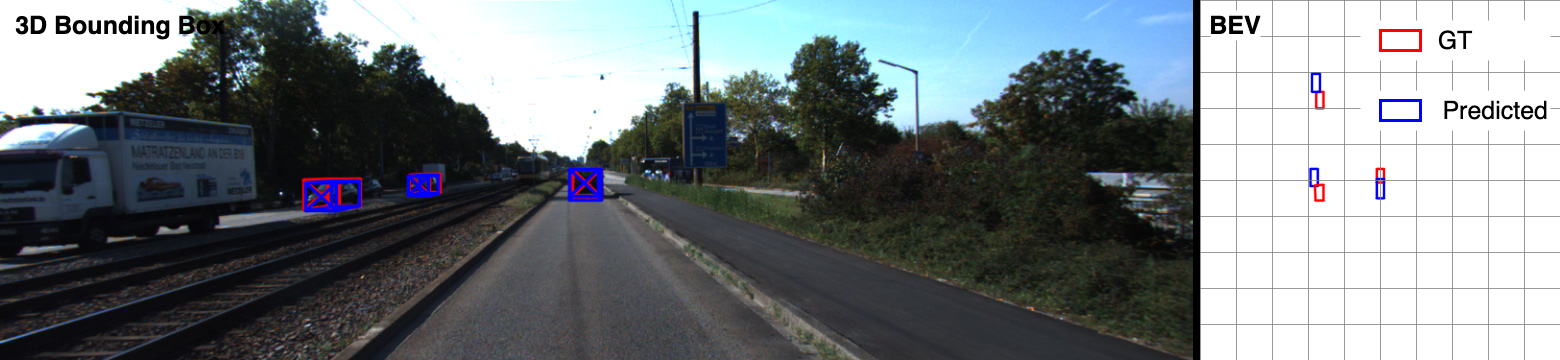}
  \end{minipage}
  \begin{minipage}[b]{0.44\linewidth}
    \centering
    \includegraphics[width=1.0\linewidth]{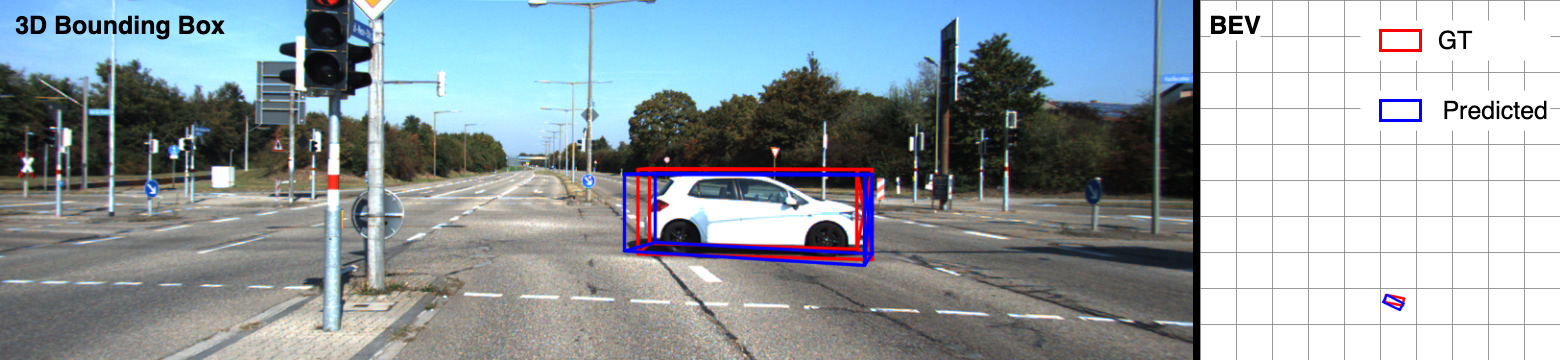}
  \end{minipage}
  \begin{minipage}[b]{0.12\linewidth}
    \centering
    \subcaption{VSRD (50\%)}
    \vspace{5mm}
  \end{minipage}
  \begin{minipage}[b]{0.44\linewidth}
    \centering
    \includegraphics[width=1.0\linewidth]{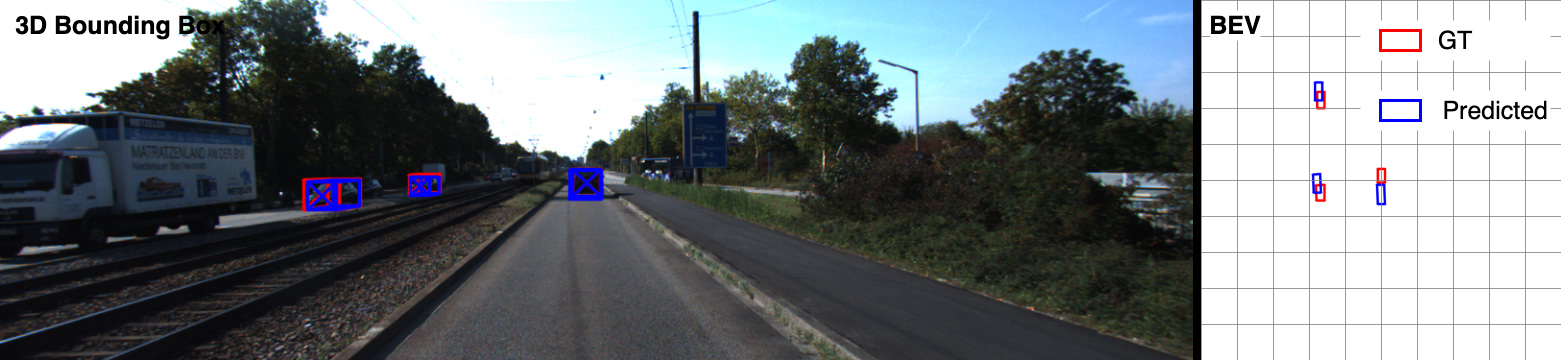}
  \end{minipage}
  \begin{minipage}[b]{0.44\linewidth}
    \centering
    \includegraphics[width=1.0\linewidth]{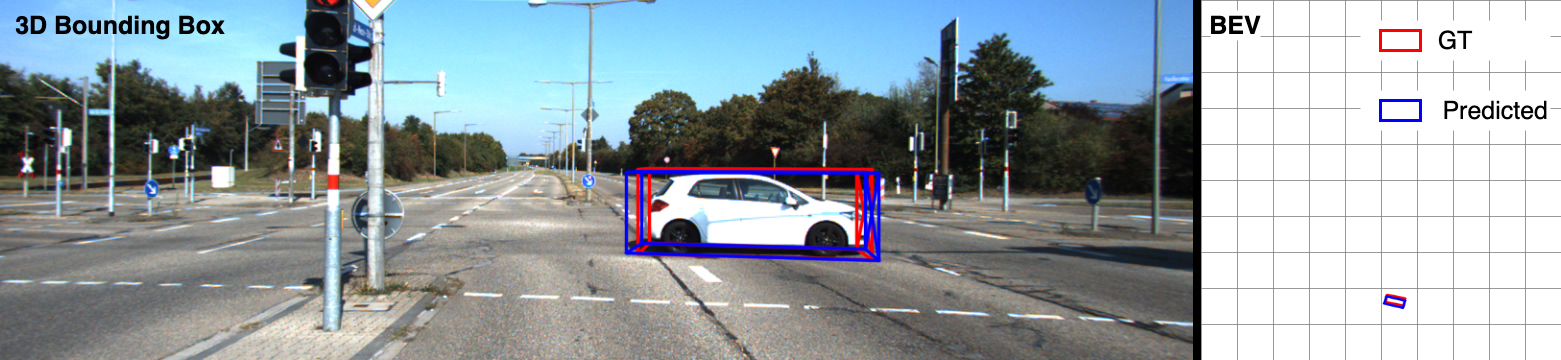}
  \end{minipage}
  \begin{minipage}[b]{0.12\linewidth}
    \centering
    \subcaption{VSRD (75\%)}
    \vspace{5mm}
  \end{minipage}
  \begin{minipage}[b]{0.44\linewidth}
    \centering
    \includegraphics[width=1.0\linewidth]{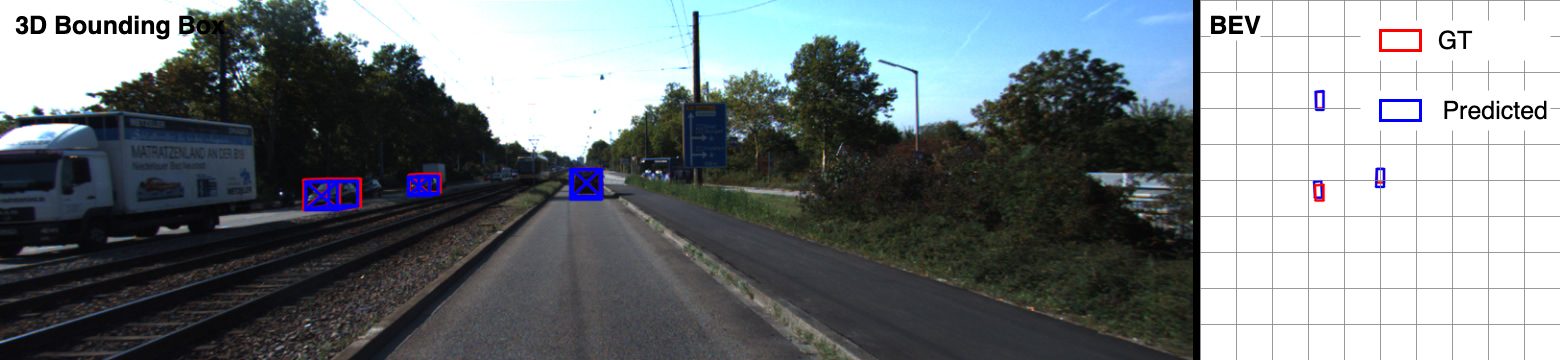}
  \end{minipage}
  \begin{minipage}[b]{0.44\linewidth}
    \centering
    \includegraphics[width=1.0\linewidth]{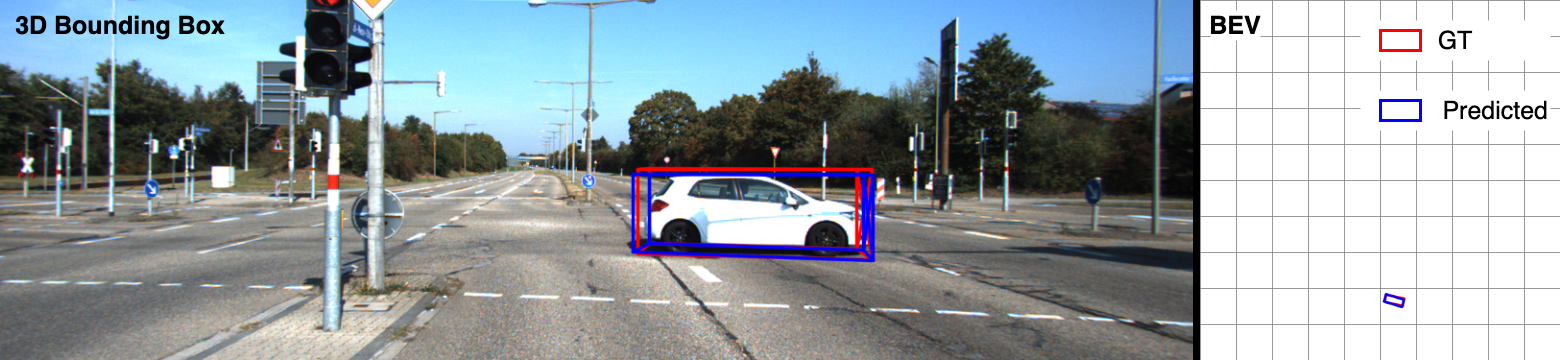}
  \end{minipage}
  \caption{Visualization results of semi-supervised monocular 3D object detection compared with MonoFlex \cite{MonoFlex}. The ground truth and predicted bounding boxes are drawn in \textcolor{red}{red} and \textcolor{blue}{blue}, respectively.}
  \label{fig:additional_visualization_results/monocular_3d_object_detection/semi_supervised_setting/monoflex}
\end{figure*}